\documentclass{sigkddExp}
\begin{document}
%
% --- Author Metadata here ---
% -- Can be completely blank or contain 'commented' information like this...
%\conferenceinfo{WOODSTOCK}{'97 El Paso, Texas USA} % If you happen to know the conference location etc.
%\CopyrightYear{2001} % Allows a non-default  copyright year  to be 'entered' - IF NEED BE.
%\crdata{0-12345-67-8/90/01}  % Allows non-default copyright data to be 'entered' - IF NEED BE.
% --- End of author Metadata ---

\title{Federated Graph Machine Learning: A Survey of Concepts, Techniques, and Applications}
%\subtitle{[Extended Abstract]
% You need the command \numberofauthors to handle the "boxing"
% and alignment of the authors under the title, and to add
% a section for authors number 4 through n.
%
% Up to the first three authors are aligned under the title;
% use the \alignauthor commands below to handle those names
% and affiliations. Add names, affiliations, addresses for
% additional authors as the argument to \additionalauthors;
% these will be set for you without further effort on your
% part as the last section in the body of your article BEFORE
% References or any Appendices.

\numberofauthors{1}
%
% You can go ahead and credit authors number 4+ here;
% their names will appear in a section called
% "Additional Authors" just before the Appendices
% (if there are any) or Bibliography (if there
% aren't)

% Put no more than the first THREE authors in the \author command
%%You are free to format the authors in alternate ways if you have more 
%%than three authors.

\author{
%
% The command \alignauthor (no curly braces needed) should
% precede each author name, affiliation/snail-mail address and
% e-mail address. Additionally, tag each line of
% affiliation/address with \affaddr, and tag the
%% e-mail address with \email.
\alignauthor Xingbo Fu, Binchi Zhang, Yushun Dong, Chen Chen, Jundong Li \\
University of Virginia\\
\{xf3av, epb6gw, yd6eb, zrh6du, jundong\}@virginia.edu
%       \affaddr{University of Virginia}\\
%       \affaddr{Address}\\
%       \affaddr{City, Country}\\
%       \email{\{xf3av, epb6gw, yd6eb, zrh6du, jundong\}@virginia.edu}

}

\maketitle
\begin{abstract}
Graph machine learning has gained great attention in both academia and industry recently. Most of the graph machine learning models, such as Graph Neural Networks (GNNs), are trained over massive graph data. However, in many real-world scenarios, such as hospitalization prediction in healthcare systems, the graph data is usually stored at multiple data owners and cannot be directly accessed by any other parties due to privacy concerns and regulation restrictions. Federated Graph Machine Learning (FGML) is a promising solution to tackle this challenge by training graph machine learning models in a federated manner. In this survey, we conduct a comprehensive review of the literature in FGML. Specifically, we first provide a new taxonomy to divide the existing problems in FGML into two settings, namely, \emph{FL with structured data} and \emph{structured FL}. Then, we review the mainstream techniques in each setting and elaborate on how they address the challenges under FGML. In addition, we summarize the real-world applications of FGML from different domains and introduce open graph datasets and platforms adopted in FGML. Finally, we present several limitations in the existing studies with promising research directions in this field.\\
\end{abstract}

\section{Introduction}
In recent years, graphs have been widely used to represent complex data in a wide diversity of real-world domains, e.g., healthcare \cite{panagopoulos2021healthcare2, wang202healthcare1}, transportation \cite{li2021traffic1, zhang2021traffic2}, bioinformatics \cite{mahmood2021bioinformatics1, zhang2021bioinformatics2}, and recommendation systems \cite{chen2021rs1, gao2022rs2}. Numerous graph machine learning techniques provide insights into understanding rich information hidden in graphs and show expressive performance in different tasks, such as node classification \cite{hang2021node_class2, zhao2021node_class1} and link prediction \cite{cai2021link_pred1, daza2021link_pred2}.

Although these graph machine learning techniques have made significant progress, most of them require a massive amount of graph data centrally stored on a single machine. However, with the emphasis on data security and user privacy \cite{voigt2017gdpr}, this requirement is often infeasible in the real world. Instead, graph data is usually distributed in multiple data owners (i.e., data isolation) and we are not able to collect the graph data in different places due to privacy concerns. For instance, a third-party company aims to train a graph machine learning model for a group of financial institutions to help them detect potential financial crimes and fraud customers. Each financial institution owns its local dataset of customers, such as their demographics, as well as transaction records among them. The customers in each financial institution form a customer graph where edges represent the transaction records. Due to strict privacy policies and commercial competition, local customer data in each institution cannot be directly shared with the company or other institutions. Meanwhile, some institutions may have connections with others, which could be viewed as structural information among institutions. Generally, the major challenge for the company lies in training a graph machine learning model for financial crime detection based on the local customer graphs and structural information among institutions without directly accessing local customer data in each institution.

Federated Learning (FL) \cite{mcmahan2017fl} is a distributed learning scheme which addresses the data isolation problem through collaborative training. It enables participants (i.e., clients) to jointly train a machine learning model without sharing their private data. Therefore, combining FL with graph machine learning becomes a promising solution to the aforementioned problem. In this paper, we term it Federated Graph Machine Learning (FGML). In general, FGML can be categorized as two settings with respect to the level of structural information. The first setting is \textit{FL with structured data}. In FL with structured data, clients collaboratively train a graph machine learning model based on their graph data while keeping the graph data locally. The second setting is \textit{structured FL}. In structured FL, there are structural information among the clients which forms a client-level graph. The client graph could be leveraged to design more effective federated optimization approaches.

While FGML provides a promising paradigm, the following challenges emerge and need to be addressed.

\begin{enumerate}
    \item \emph{Cross-client missing information}. A common scenario in FL with structured data is that each client owns a subgraph of the global graph and some nodes may have neighbors belonging to other clients. Due to privacy concerns, a node can only aggregate the features of its neighbors within the client but cannot access the features of those located on other clients, which leads to insufficient node representations \cite{chen2021fedgraph, peng2021fedni, zhang2021ppsgcn, zhang2021fedsage}.
    \item \emph{Privacy leakage of graph structures}. In traditional FL, a client is not allowed to expose the features and labels of its data samples. In Fl with structured data, the privacy of structural information should also be considered. The structural information can be either directly exposed by sharing the adjacency matrix or indirectly exposed by transmitting node embeddings \cite{liu2021fesog, ru2021cdp, wu2022fedpergnn, zhang2022fedr}.
    \item \emph{Data heterogeneity across clients}. Unlike traditional FL where data heterogeneity comes from non-IID %features of 
    data samples \cite{huang2021fedamp, shamsian2021pflhn}, graph data in FGML contains rich structural information \cite{jin2020structure, johnson2020structure, liu2022structure, zhao2021data_aug}. Meanwhile, divergent graph structures across clients can also affect performance of graph machine learning models.
    \item \emph{Parameter utilization strategies}. In structured FL, the client graph enables a client to obtain information from its neighbor clients. The effective strategies of fully utilizing neighbor information orchestrated by a central server or in a fully decentralized manner should be well designed in structured FL \cite{he2021spreadgnn, lalitha2019p2p, meng2021cnfgnn}.
    
\end{enumerate}

To tackle the above challenges, a great number of algorithms have been proposed in recent years. However, to the best of our knowledge, the existing surveys mainly focus on challenges and approaches in standard FL \cite{kairouz2021fl_survey2, li2020fl_survey1, tan2022fl_survey5, yang2019fl_survey3, zhu2021fl_survey4} yet only a few attempts have been made to survey specific problems and techniques in FGML \cite{liu2022review2, zhang2021review1}. A position paper \cite{zhang2021review1} provides a categorization of FGML but does not summarize main techniques in FGML. Another review paper \cite{liu2022review2} only covers a limited number of related papers in this topic and introduces the existing techniques very briefly.

In this survey, we introduce the concepts of two problem settings in FGML. Then we review the current techniques under each setting and introduce real-world applications in FGML. We also summarize accessible graph datasets and platforms which can be used for applications of FGML. Finally, several promising future directions are presented. Our contributions in this paper can be summarized as follows.

\begin{itemize}
    \item \textbf{Taxonomy of Techniques in FGML.} We propose a taxonomy of FGML based on different problem settings and summarize key challenges in each setting.
    \item \textbf{Comprehensive Technique Review.} We provide a comprehensive overview of the existing techniques in FGML. Compared with the existing reviews, we not only investigate a more extensive set of related work but also provide a more elaborate analysis of techniques instead of simply listing the steps of each method.
    \item \textbf{Real-World Applications.} We are the first to summarize real-world applications of FGML. We categorize the applications by their domains and introduce related works in each domain.
    \item \textbf{Datasets and Platforms.} We introduce the existing datasets and platforms in FGML, which facilitates developing algorithms and deploying applications in FGML for engineers and researchers.
    \item \textbf{Promising Future Directions.} We point out the limitations of the existing methods and provide promising future directions in FGML.
\end{itemize}

The rest of this paper is organized as follows. Section 2 briefly introduces definitions in graph machine learning as well as concepts and challenges of two settings in FGML. We review mainstream techniques in the two settings in Section 3 and Section 4, respectively. Section 5 further explores applications of FGML in the real world. Section 6 presents open graph datasets used in related FGML papers and two platforms for FGML. We also provide possible future directions in Section 7. Finally, Section 8 concludes this paper.

\section{Problem Formulation}
In this section, we first present related definitions in graph machine learning and FL. Then we introduce the problem formulation of two different settings in FGML.

\noindent \textbf{Notations.} Throughout this paper, we use bold lowercase letters (e.g., $\textbf{z}$) and bold uppercase letters (e.g., $\textbf{A}$) to represent vectors and matrices, respectively. For any matrix, e.g., $\textbf{A}$, we use $\textbf{A}_i$ to denote its $i$-th row vector and $\textbf{A}_{ij}$ to denote its $(i,j)$-th entry. The $l_p$ norm of a vector $\textbf{z}$ for $p \geq 1$ is denoted as $||\textbf{z}||_p$. We use letters in calligraphy font (e.g., $\mathcal{V}$) to denote sets. $|\mathcal{V}|$ denotes the cardinality of set $\mathcal{V}$.

\subsection{Graph Machine Learning}

\textbf{Definition 1. (Graphs)} \textit{A graph is $\mathcal{G}=(\mathcal{V}, \mathcal{E})$, where $\mathcal{V}$ is the node set and $\mathcal{E}$ is the edge set. $v_i\in \mathcal{V}$ denotes a node and $e_{ij}=(v_i, v_j)\in \mathcal{E}$ denotes an edge between node $v_i$ and node $v_j$.}\\

We use $\textbf{A}\in \{0, 1\}^{n\times n}$ to represent the adjacency matrix of graph $\mathcal{G}$, where $n=|\mathcal{V}|$ is the total number of nodes. $\textbf{A}_{ij}=1$ implies that there exists an edge between node $v_i$ and node $v_j$, otherwise $\textbf{A}_{ij}=0$. $\textbf{D}\in \mathbb{R}^{n\times n}$ denotes the degree diagonal matrix where $\textbf{D}_{ii}=\sum_j \textbf{A}_{ij}$. 
%$\textbf{L}=(\textbf{D}+\textbf{I})^{-\frac{1}{2}}(\textbf{A}+\textbf{I})(\textbf{D}+\textbf{I})^{-\frac{1}{2}}$ denotes the \textcolor{red}{normalized adjacency matrix}. 
The neighborhood of node $v_i$ is defined as $N(v_i)=\{v_j\in\mathcal{V}|(v_i, v_j)\in \mathcal{E}\}$. For graph data with node features, we use $\textbf{X}\in \mathbb{R}^{n\times d}$ to denote the node feature matrix where $d$ is the number of node features.

Graphs can be categorized as homogeneous graphs (containing only one type of nodes and one type of edges) and heterogeneous graphs (whose nodes belong to more than one type of nodes and/or edges) according to the number of node types and edge types. The two typical heterogeneous graphs we mention in this paper are knowledge graphs (KGs) and user-item graphs.\\

%Considering the number of node types and edge types, graphs can be categorized as homogeneous graphs (containing only one type of nodes and one type of edges) and heterogeneous graphs (whose nodes belong to more than one type of nodes and/or edges). The two typical heterogeneous graphs we mention in this paper are knowledge graphs (KGs) and user-item graphs.\\

\textbf{Definition 2. (Knowledge Graphs)} \textit{A knowledge graph is a directed heterogeneous graph $\mathcal{G}=(\mathcal{V}, \mathcal{E})$ where nodes are entities and edges are subject-property-object triple facts. Each edge of the form (head entity, relation, tail entity) (denoted as $(h,r,t)$) indicates a relationship $r$ from a head entity $h$ to a tail entity $t$.}\\

\textbf{Definition 3. (User-Item Graphs)} \textit{A user-item graph is a heterogeneous graph $\mathcal{G}=(\mathcal{V}, \mathcal{E})$. Users and items serve as nodes and relations between users and items serve as edges. In some scenarios, relations also exist between users and between items.}\\

\begin{figure*}
\centering
\includegraphics[width=\linewidth]{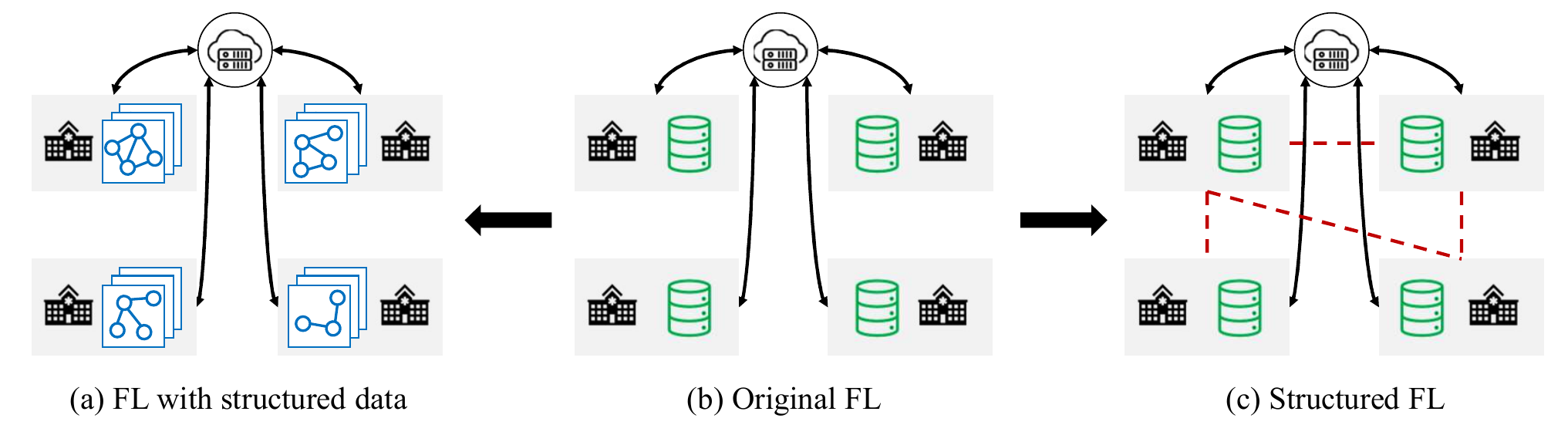}
\caption{The framework comparison among original FL, FL with structured data and structured FL. (a) FL with structured data: each client owns graph data, i.e., a (sub)graph or multiple graphs. (b) Original FL: data samples on clients are typically Euclidean data and no links exist among edges. (c) Structured FL: clients are connected by links and form a client graph.}
\label{fig:fl}
\end{figure*}

\textbf{Definition 4. (Graph Machine Learning Models)} \textit{Given a graph $\mathcal{G}=(\mathcal{V}, \mathcal{E})$, a graph machine learning model $f_\omega$ parameterized by $\omega$ learns the node representations} $\textbf{H} \in \mathbb{R}^{n\times d_e}$ \textit{with respect to $\mathcal{G}$ for downstream tasks, where $d_e$ is the dimension of node embeddings}
\begin{equation} \label{gcn}
   \textbf{H} = f_\omega(\mathcal{G}).
\end{equation}

For node classification tasks, we usually employ a softmax function to obtain the probability vector for each node based on its embedding, and then a loss function (e.g., cross entropy) is applied to measure the difference between predictions and the given node labels.

For graph classification tasks, a graph-level representation $\textbf{h}_{\mathcal{G}}$ can be pooled from node representations
\begin{equation} \label{readout}
    \textbf{h}_{\mathcal{G}} = \text{readout}(\textbf{H}),
\end{equation}
where $\text{readout}(\cdot)$ is a pooling function (e.g., mean pooling and sum pooling) which aggregates the embeddings of all nodes in the graph into a single embedding vector.
%where $\text{readout}(\cdot)$ can be implemented as mean pooling, sum pooling, etc, which essentially aggregates the embeddings of all nodes on the graph into a single embedding vector.

Without loss of generality, we mainly consider Graph Neural Networks (GNNs) (e.g., GCN \cite{kipf2016gcn}, GAT \cite{velivckovic2017gat}, and GraphSage \cite{hamilton2017graphsage}) as graph machine learning models in this survey. In GNNs, each node $v_i$ typically gathers the information from its neighbors $N(v_i)$ and aggregates them with its own information to update its representation $\textbf{h}_i$. Mathematically, an $L$-layer GNN $f_\omega$ can be formulated as 
\begin{equation}
    \textbf{h}_i^l = \sigma (\omega^l \cdot (\textbf{h}_i^{l-1}, \text{Agg}(\{\textbf{h}_j^{l-1}|v_j \in N(v_i) \})))
\end{equation}
for $l=1,2,\cdots, L$, where $\textbf{h}_i^l$ is the representation of node $v_i$ after the $l$-layer of $f_\omega$ and $\textbf{h}_i^0=\textbf{X}_i$ is the raw feature of node $v_i$. $\omega^l$ is the learnable parameters in the $l$-layer of $f_\omega$ and $\text{Agg}(\cdot)$ is the aggregation operation (e.g., mean pooling). $\sigma$ is an activation function.

\subsection{Federated Learning}
FGML is a type of FL which involves structural information. Before introducing the concepts of two settings in FGML, we provide the definition of FL in this subsection.\\

\textbf{Definition 5. (Federated Learning)} \textit{In standard FL, we consider a set of $M$ clients $\mathcal{C}=\{c_k\}_{k=1}^M$. Each client $c_k$ owns its private dataset $\mathcal{D}_k=\{(\textbf{x}_i, y_i)_{i=1}^{N_k}\}$ sampled from its own data distribution, where $\textbf{x}_i$ is the feature vector of $i$-th data sample and $y_i$ is the corresponding label of the data sample. $N_k=|\mathcal{D}_k|$ is the number of data samples on client $c_k$ and $N=\sum_{k=1}^M N_k$. Let $l_k$ denote the loss function parameterized by $\omega$ on client $c_k$. The goal of FL is to optimize the overall objective function while keeping private datasets locally}
\begin{equation}  \label{fl}
    \min\limits_{\omega}\sum_{k=1}^{M} \frac{N_k}{N} L_k(\omega) = \min\limits_{\omega}\frac{1}{N}\sum_{k=1}^{M}\sum_{i=1}^{N_k}l_k(\textbf{x}_i, y_i; \omega),
\end{equation}
\textit{where $L_k$ is the average loss over the local data on client $c_k$.}\\

FedAvg \cite{mcmahan2017fl} is a typical algorithm for federated optimization to obtain high model utility while preserving privacy. In FedAvg, only model parameters are transmitted between the central server and each client. Specifically, during each round $t$, the central server selects a subset of clients and sends them a copy of the current global model parameters $\omega^t$ for local training. Each selected client $c_k$ updates the received copy $\omega^t_k$ by an optimizer such as stochastic gradient descent (SGD) for a variable number of iterations locally on its own dataset $\mathcal{D}_k$. Then the server collects updated model parameters $\omega^t_k$ from the selected clients and aggregates them to obtain a new global model $\omega^{t+1}$. Finally, the server broadcasts the updated global model $\omega^{t+1}$ to clients for training in round $t+1$.

It is worthwhile to note that GNN and FL both involve an \textit{aggregation} operation. Aggregation in the context of GNN represents the operation that a node updates its representation by aggregating information from its neighbors. Aggregation in the context of FL represents the operation that the central server collects model parameters from clients and updates the global model parameters. Following the previous survey \cite{liu2022review2}, we use \textit{GNN aggregation} and \textit{FL aggregation} in this survey to represent two aggregation operations in GNN and FL, respectively.

\subsection{Federated Graph Machine Learning}
Standard FL mainly deals with tasks on Euclidean data (e.g., image classification) and equally aggregates model parameters from clients. Different from standard FL, federated graph machine learning involves structural information in federated optimization. Based on the level of structural information, FGML can be categorized as two mainstreams.

\noindent \textbf{Setting 1. (FL with Structured Data)} In FL with structured data, clients possess private structured datasets (i.e., graphs) and jointly train a graph machine learning model orchestrated by the central server. Fig.~\ref{fig:fl}(a) illustrates the framework of FL with structured data. Formally, each client $c_k$ owns its private local data $\mathcal{D}_k=\{\mathcal{G}_1, \mathcal{G}_2, \cdots\}$, where each $\mathcal{G}_i=(\mathcal{V}_i, \mathcal{E}_i)$ is a graph with its node set $\mathcal{V}_i$ and edge set $\mathcal{E}_i$. The objective of FL with structured data is that each client collaboratively trains a graph machine learning model $f_\omega$ with other clients based on its local graph dataset $\mathcal{D}_k$ while always keeping $\mathcal{D}_k$ locally. Note that each client in FL with structured data may have one single (sub)graph or multiple graphs. In general, clients train a graph machine learning model for graph-level tasks when each client $c_k$ owns multiple graphs and $N_k$ is the number of graphs on client $c_k$; on the contrary, when each client $c_k$ owns one single graph $\mathcal{G}_k$ or a subgraph $\mathcal{G}_k$ of an entire graph, the graph machine learning model is for node-level tasks and $N_k = |\mathcal{V}_k|$ is the number of nodes in $\mathcal{G}_k$.

\noindent \textbf{Setting 2. (Structured FL)} In structured FL, relations exist among clients. When we take each client as a node, all the clients in structured FL will form a graph $\mathcal{G}^\mathcal{C}=\{\mathcal{V}^\mathcal{C}, \mathcal{E}^\mathcal{C}\}$ where $\mathcal{V}^\mathcal{C}$ is the client set and $\mathcal{E}^\mathcal{C}$ contains links between clients. Fig.~\ref{fig:fl}(c) shows the framework of structured FL. Formally, given the client graph $\mathcal{G}^\mathcal{C} =\{\mathcal{V}^\mathcal{C}, \mathcal{E}^\mathcal{C}\}$, each client $c_k$ collaboratively trains a machine learning model $f_\omega$ by interacting with its neighbors $N(c_k)= \{c_s|(c_k, c_s) \in \mathcal{E}^\mathcal{C}\}$. It is noteworthy that the datasets on clients do not have to be structured data.

In Section 3 and Section 4, we review the existing techniques in FL with structured data and structured FL and analyze how they solve the aforementioned challenges, respectively. We summarize these techniques in Table \ref{table:tech}.

\begin{table*}[h]
\centering
\caption{Summary of techniques in FGML.}
\begin{tabular}{c|cccc} \hline
Settings & Techniques & Approaches & Data on each client & Downstream Tasks \\ \hline
\multirow{\NumRefFLWithStructuredData}{*}{\rotatebox{90}{FL with Structured Data}} & Cross-Client Info. Recon. & FedGraph \cite{chen2021fedgraph} & A subgraph & Node classification\\ \cline{2-5}
\multirow{\NumRefFLWithStructuredData}{*}{} & Cross-Client Info. Recon. &  Glint \cite{liu2021glint} & A subgraph & Node classification \\ \cline{2-5}
\multirow{\NumRefFLWithStructuredData}{*}{} & Cross-Client Info. Recon. & PPSGCN \cite{zhang2021ppsgcn}  & A subgraph & Node classification \\ \cline{2-5}
\multirow{\NumRefFLWithStructuredData}{*}{} & Cross-Client Info. Recon. & FedSage+ \cite{zhang2021fedsage}  & A subgraph & Node classification \\ \cline{2-5}
\multirow{\NumRefFLWithStructuredData}{*}{} & Cross-Client Info. Recon. & FedNI \cite{peng2021fedni} & A subgraph & Node classification \\ \cline{2-5}
\multirow{\NumRefFLWithStructuredData}{*}{} & Overlapping Ins. Align. & FedVGCN \cite{ni2021fedvgcn} & A graph & Node classification \\ \cline{2-5}
\multirow{\NumRefFLWithStructuredData}{*}{} & Overlapping Ins. Align. & VFGNN \cite{zhou2020vfgnn} & A graph & Node classification \\ \cline{2-5}
\multirow{\NumRefFLWithStructuredData}{*}{} & Overlapping Ins. Align. & FedSGC \cite{cheung2021fedsgc} & A graph & Node classification \\ \cline{2-5}
\multirow{\NumRefFLWithStructuredData}{*}{} & Overlapping Ins. Align. & SGNN \cite{mei2019sgnn} & A graph & Node classification \\ \cline{2-5}
\multirow{\NumRefFLWithStructuredData}{*}{} & Overlapping Ins. Align. & FedGL \cite{chen2021fedgl} & A subgraph & Node classification \\ \cline{2-5}
\multirow{\NumRefFLWithStructuredData}{*}{} & Overlapping Ins. Align. & FedE\cite{chen2021fede} & A KG & KG completion \\ \cline{2-5}
\multirow{\NumRefFLWithStructuredData}{*}{} & Overlapping Ins. Align. & FedR \cite{zhang2022fedr} & A KG & KG completion \\ \cline{2-5}
\multirow{\NumRefFLWithStructuredData}{*}{} & Overlapping Ins. Align. & FKGE \cite{peng2021fkge} & A KG & KG completion \\ \cline{2-5}
\multirow{\NumRefFLWithStructuredData}{*}{} & Overlapping Ins. Align. & FedGNN \cite{wu2021fedgnn} & A user-item graph & Rating prediction \\ \cline{2-5}
\multirow{\NumRefFLWithStructuredData}{*}{} & Overlapping Ins. Align. & FedPerGNN \cite{wu2022fedpergnn} & A user-item graph & Rating prediction \\ \cline{2-5}
\multirow{\NumRefFLWithStructuredData}{*}{} & Overlapping Ins. Align. & FeSoG \cite{liu2021fesog} & A user-item graph & Rating prediction \\ \cline{2-5}
\multirow{\NumRefFLWithStructuredData}{*}{} & Non-IID data adaptation & FedAlign \cite{lin2020fedalign}  & A subgraph & Node classification \\ \cline{2-5}
\multirow{\NumRefFLWithStructuredData}{*}{} & Non-IID data adaptation & FLIT+ \cite{zhu2021filt} & Multiple graphs & Graph classification/regression \\ \cline{2-5}
\multirow{\NumRefFLWithStructuredData}{*}{} & Non-IID data adaptation & GraphFL \cite{wang2020graphfl}  & A subgraph & Node classification \\ \cline{2-5}
\multirow{\NumRefFLWithStructuredData}{*}{} & Non-IID data adaptation & FML-ST \cite{li2022fml_st} & A graph & Node regression \\ \cline{2-5}
\multirow{\NumRefFLWithStructuredData}{*}{} & Non-IID data adaptation & FedGCN \cite{hu2022fedgcn} & Multiple graphs & Graph classification \\ \cline{2-5}

\multirow{\NumRefFLWithStructuredData}{*}{} & Non-IID data adaptation & ASFGNN \cite{zheng2021asfgnn} & A subgraph & Node classification \\ \cline{2-5}

\multirow{\NumRefFLWithStructuredData}{*}{} & Non-IID data adaptation & GCFL+ \cite{xie2021gcfl} & Multiple graphs & Graph classification \\ \cline{2-5}
\multirow{\NumRefFLWithStructuredData}{*}{} & Non-IID data adaptation & CTFL \cite{zhang2022ctfl} & A subgraph & Node regression    \\ \hline

\multirow{\NumRefStructuredFL}{*}{\rotatebox{90}{Structured FL}} & Centralized Aggregation & SFL \cite{chen2022sfl}  & A node & Time series prediction \\ \cline{2-5}
\multirow{\NumRefStructuredFL}{*}{} & Centralized Aggregation & BiG-Fed \cite{xingbigfed} & A node & Time series prediction  \\ \cline{2-5}
\multirow{\NumRefStructuredFL}{*}{} & Centralized Aggregation & CNFGNN \cite{meng2021cnfgnn} & A node & Time series prediction \\ \cline{2-5}

\multirow{\NumRefStructuredFL}{*}{} & Fully Decentralized Trans. & D-FedGNN \cite{pei2021dfedgnn} & A node & Graph classification/regression \\ \cline{2-5}
\multirow{\NumRefStructuredFL}{*}{} & Fully Decentralized Trans. & \cite{lalitha2019p2p}  & A node & Node regression \\ \cline{2-5}
\multirow{\NumRefStructuredFL}{*}{} & Fully Decentralized Trans. & GFL \cite{rizk2021gfl}  & A node & Node regression \\ \cline{2-5}
\multirow{\NumRefStructuredFL}{*}{} & Fully Decentralized Trans. & FL DSGD/DSGT \cite{lu2020dsgd2} & A node & Health record representation \\ \cline{2-5}
\multirow{\NumRefStructuredFL}{*}{} & Fully Decentralized Trans. & FD DSGD/DSGT \cite{lu2019dsgd3} & A node & Health record representation \\ \cline{2-5}
\multirow{\NumRefStructuredFL}{*}{} & Fully Decentralized Trans. & \cite{xing2020dsgd} & A node & Image classification \\ \cline{2-5}
\multirow{\NumRefStructuredFL}{*}{} & Fully Decentralized Trans. & SpreadGNN \cite{he2021spreadgnn} & A node & Graph classification/regression  \\ \cline{2-5}
\multirow{\NumRefStructuredFL}{*}{} & Fully Decentralized Trans. & cPDS \cite{brisimi2018health} & A node & Node classification \\ \cline{2-5}
\multirow{\NumRefStructuredFL}{*}{} & Fully Decentralized Trans. & dFedU \cite{dinh2021dfedu} & A node & Image classification\\ \cline{2-5}
\multirow{\NumRefStructuredFL}{*}{} & Fully Decentralized Trans. & EF-HC \cite{zehtabi2022efhc} & A node & Image classification
\\  \hline\end{tabular}
\label{table:tech}
\end{table*}

\section{FL with Structured Data}
\begin{figure}
\centering
\includegraphics[width=\linewidth]{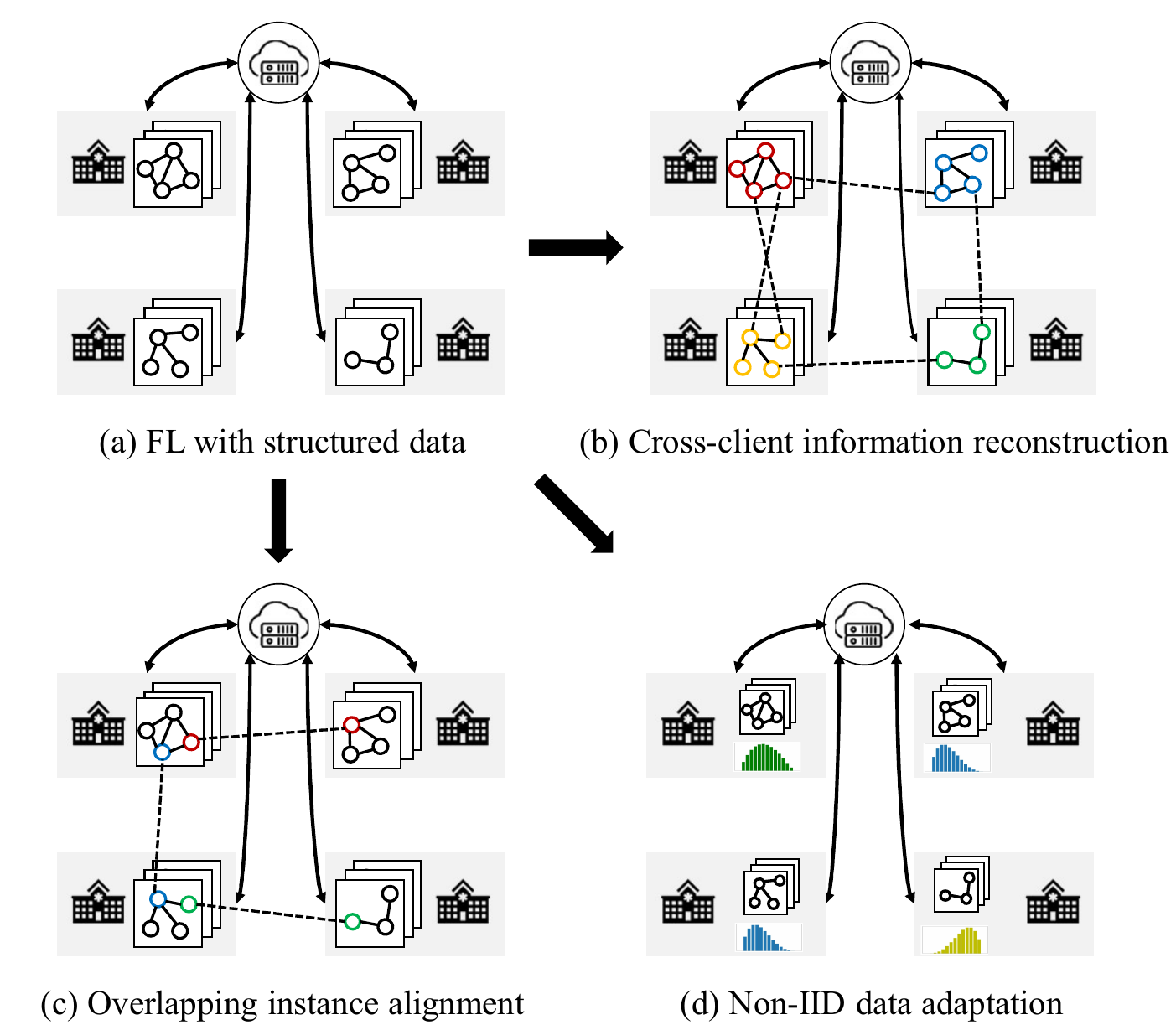}
\caption{The taxonomy of techniques in FL with structured data. The techniques in (a) FL with structured data can be categorized as (b) cross-client information reconstruction which recovers missing links (e.g., dashed lines in (b)) between nodes from different clients, (c) overlapping instance alignment which aligns overlapping instances (e.g., nodes connected by dashed lines in (c)) among different clients,  and (d) Non-IID data adaptation which tackles the non-IID characteristic of data across clients.}
\label{fig:structure3}
\end{figure}

The goal of clients in FL with structured data is to jointly train a graph machine learning model based on their local graph datasets while preserving privacy. In this section, we review techniques in FL with structured data for improving model utility and tackling the aforementioned challenges. Fig.~\ref{fig:structure3} illustrates the taxonomy of techniques in FL with structured data.

\subsection{Cross-Client Information Reconstruction}
When a graph is split into multiple subgraphs and each client owns a subgraph of the original graph, each node can only perform GNN aggregation on the information from a subset of its neighbors (i.e., those within the subgraph) but cannot obtain information from those located on other clients due to the privacy issue. The missing cross-client information leads to biased node embeddings on each client and therefore degrades the performance of graph machine learning models. The objective in this case is to reconstruct the important missing cross-client information for calculating node embeddings. The existing techniques can be categorized as intermediate result transmission and missing neighbor generation. The difference lies in whether the original global graph structure is known: the studies in intermediate result transmission assume that the central server is aware of the original graph structure, while those in missing neighbor generation do not.

\subsubsection{Intermediate Result Transmission}
When the central server is aware of the original graph structure (including missing cross-client links), it is able to collect intermediate results (e.g., node representations) in graph machine learning models from clients and compute node embeddings according to their complete neighbor lists including cross-client neighbors.

Considering an $L$-layer GCN model $f_\omega$, the operation in the $l$-th layer of $f_\omega$ can be written as
\begin{equation}\label{eq:vanilla_gcn}
    \textbf{H}^l=\sigma\left(\textbf{LH}^{l-1}\textbf{W}^l\right),
\end{equation}
where $\textbf{H}^l$ is the hidden representation after the $l$-th layer of $f_\omega$ and $\textbf{H}^0=\textbf{X}$. $\textbf{W}^l$ denotes the weight matrix of the $l$-th layer and $\textbf{L}=(\textbf{D}+\textbf{I})^{-\frac{1}{2}}(\textbf{A}+\textbf{I})(\textbf{D}+\textbf{I})^{-\frac{1}{2}}$.
In the federated setting, we rewrite the vanilla GCN model in a distributed manner, where the hidden matrix $\textbf{H}^l_{(k)}$ after the $l$-th layer of $f_\omega$ on each client $c_k$ can be computed by
\begin{equation} \label{fed_gcn_layer}
\begin{aligned}
    \textbf{H}^l_{(k)} &=\sigma\left(\sum_{s=1}^M \textbf{L}_{(ks)}\textbf{H}^{l-1}_{(s)}\textbf{W}^l_{(s)}\right)\\
    &=\sigma\left(\textbf{L}_{(kk)}\textbf{H}^{l-1}_{(k)}\textbf{W}^l_{(k)}+\sum_{s\neq k} \textbf{L}_{(ks)}\textbf{H}^{l-1}_{(s)}\textbf{W}^l_{(s)}\right)
\end{aligned}
\end{equation}
for $l=1,2,\cdots, L$, where $\textbf{H}^0_{(k)}=\textbf{X}_{(k)}$ are node features on client $c_k$. $\textbf{W}^l_{(k)}$ denotes the local parameters in the $l$-th layer of the local graph machine learning model on client $c_k$. $\textbf{L}_{(ks)}$ is a block of $\textbf{L}$ corresponding to the rows of nodes in $\mathcal{V}_k$ and columns of nodes in $\mathcal{V}_s$.

An intuitive way of obtaining information from a node's neighbors located on other clients is to transmit node embeddings directly. To avoid exposing raw node features, each client performs GNN aggregation within its local subgraph when processing the first GCN layer \cite{chen2021fedgraph, liu2021glint}
\begin{equation} \label{fedgraph}
    \textbf{H}^1_{(k)} =\sigma(\textbf{L}_{(kk)}\textbf{X}_{(k)}\textbf{W}^1_{(k)}).
\end{equation}
Then, each client performs GNN aggregation for a node including the embeddings of its neighbors from other clients following Eq.~(\ref{fed_gcn_layer}) after the first GCN layer.

However, the block of degree diagonal matrix $\textbf{D}_{(k)} \in \mathbb{R}^{N_k \times N_k}$ containing node degree information on client $c_k$ is unknown for other clients due to privacy concerns and must be kept locally during computation. There have been a series of corresponding solutions to this problem. For instance, PPSGCN \cite{zhang2021ppsgcn} reformulates each block $\textbf{L}_{(ks)}$ by
\begin{equation}
    \textbf{L}_{(ks)}=(\textbf{D}_{(k)}+\textbf{I})^{-\frac{1}{2}}\cdot \tilde{\textbf{L}}_{(ks)},
\end{equation}
where $\tilde{\textbf{L}}_{(ks)}=\textbf{A}_{(ks)}(\textbf{D}_{(k)}+\textbf{I})^{-\frac{1}{2}}$. Therefore, Eq.~(\ref{fed_gcn_layer}) can be rewritten as 
\begin{equation} \label{ppsgcn}
    \begin{aligned}
        \textbf{H}^l_{(k)} =\sigma(&\textbf{L}_{(kk)}\textbf{H}^{l-1}_{(k)}\textbf{W}^l_{(k)}\\
    &+(\textbf{D}_{(k)}+\textbf{I})^{-\frac{1}{2}}\sum_{s\neq k} \tilde{\textbf{L}}_{(ks)}\textbf{H}^{l-1}_{(s)}\textbf{W}^l_{(s)})
\end{aligned}
\end{equation}
for $l=1,2,\cdots, L$, where $\tilde{\textbf{L}}_{(ks)}\textbf{H}^{l-1}_{(s)}\textbf{W}^l_{(s)}$ can be calculated locally within client $c_s$ and transmitted to the server. Consequently, we can learn node representations over clients without exchanging local graph structure information.

Although the raw data (i.e., node features and structural information) are preserved with intermediate result transmission, it requires the structural information of the original graph to compute node embeddings, which might be impractical in the real world. Furthermore, intermediate results are transmitted multiple times for each local update, which also brings significant communication costs.

\subsubsection{Missing Neighbor Generation}
When the original graph structure is unknown to the server, the techniques in intermediate result transmission fail since the cross-subgraph links carrying important information will never be captured by any client \cite{zhang2021fedsage}. To tackle this issue, several approaches of missing neighbor generation have been proposed recently. 

The intuition of missing neighbor generation is to design a missing neighbor generator to reconstruct the features of a node's cross-subgraph neighbors on other clients \cite{peng2021fedni, zhang2021fedsage}. Concretely, each client $c_k$ first hides a subset of nodes and related edges in its local subgraph $\mathcal{G}_k$ following a specific strategy (e.g., Breadth-First Search) \cite{peng2021fedni} to form an impaired subgraph $\bar{\mathcal{G}}_k$. Then each client trains $\mathcal{X}$ a predictor parameterized by $\theta_d$ for predicting the number $\tilde{n}_i$ of the masked neighbors of each node $v_i$ in $\bar{\mathcal{G}}_k$ and an encoder (e.g., GCN) parameterized by $\theta_f$ for predicting the features $\tilde{\textbf{X}}_{(i)}$ of the masked neighbors by minimizing the loss
\begin{equation} \label{fedsage}
    \begin{aligned}
            L^n = \lambda^d L^d(\tilde{n}_i, n_i;\theta_d) +\lambda^f L^f(\{\tilde{\textbf{X}}_{(i)}\}_{v_i \in \bar{\mathcal{G}}_k}, \{\textbf{X}_{(i)}\}_{v_i \in \bar{\mathcal{G}}_k};\theta_f),
    \end{aligned}
\end{equation}
where $L^d$ and $L^f$ are the loss functions for the predicted number of masked neighbors and their predicted features, respectively. $\textbf{X}_{(i)}$ is the features of node $v_i$'s masked neighbors. A cross-subgraph feature reconstruction term \cite{zhang2021fedsage} is introduced to $L^f$, aiming to recover features of cross-subgraph missing neighbors by decreasing the distance between predicted node features and the closest node feature on other clients. 
%Edges between missing nodes and known nodes can be constructed based on phenotype information \cite{peng2021fedni}.

\subsection{Overlapping Instance Alignment}
In applications, an instance (e.g., a node in homogeneous graphs or an entity in KGs) could belong to two or more clients. Under this setting, the embeddings of an overlapping instance from different clients may come from different embedding spaces during collaborative training. To tackle this problem, the overlapping instance alignment technique is proposed \cite{chen2021fedgl, peng2021fkge, zhou2020vfgnn}.
%The existing techniques about overlapping instance alignment are applied to the scenario where an instance (e.g., a node in homogeneous graphs or an entity in KGs) belonging to a client also exists on other clients \cite{chen2021fedgl, zhou2020vfgnn, peng2021fkge}. 
The key idea is to learn global instance embeddings based on the local instance embeddings from clients. This technique can be applied to homogeneous graphs, KGs, and user-item graphs.

\subsubsection{Homogeneous Graph-Based Alignment} 
The existing works about instance alignment in homogeneous graphs are mainly under vertical FL \cite{yang2019fl_survey3}. In generic vertical FL, clients have overlapping nodes but differ in the feature space. Unlike generic vertical FL, structural information in graph data is also taken into account in FGML. One common scenario is that a set of nodes are located on all clients but their features and relations are different across clients \cite{ni2021fedvgcn, zhou2020vfgnn}. The central server can collect local node embeddings from clients and align overlapping node embeddings. For instance, VFGNN \cite{zhou2020vfgnn} first computes local node embeddings $\textbf{H}_{(k)}$ on each client $c_k$ using a graph machine learning model. Then it combines the embeddings of each node $v_i$ via a combination strategy
\begin{equation} \label{vfgnn}
\textbf{H} \leftarrow \text{COMBINE}(\{\textbf{H}_{(k)}\}_{k=1}^M),
\end{equation}
where $\text{COMBINE}(\cdot)$ denotes a combination operator (e.g., Concat, Mean and Regression).

Another scenario is that one client only contains structural information of graph data and other clients only contain node features \cite{cheung2021fedsgc}. In this scenario, graph machine learning models cannot be simply applied on each client since structural information and node features are located in different clients. The related techniques deal with this problem by protecting structural information and raw node features simultaneously during federated optimization. For example, SGNN \cite{mei2019sgnn} replaces the original adjacency matrix with a structural similarity matrix $\textbf{A}^s$ where entry $\textbf{A}^s_{ij}$ measures the structural similarity between node $v_i$ and node $v_j$. SGNN computes $\textbf{A}^s_{ij}$ by
\begin{equation} \label{SGNN}
\textbf{A}^s_{ij}=\text{exp}(- \text{dist}(\text{OD}(N(v_i)), \text{OD}(N(v_j)))),
\end{equation}
where $\text{OD}(\cdot)$ returns a list of ordered degree given the input node list and $\text{dist}(\cdot,\cdot)$ is a distance function (e.g., dynamic time warping (DTW) \cite{olsen2018dtw}). Then SGNN embeds original features on the clients which only contain features using one-hot encoding and finally computes node embeddings with the structural similarity matrix. FedSGC \cite{cheung2021fedsgc} applies Homomorphic Encryption (HE) \cite{acar2018he} for secure transmission of the adjacency matrix and node features.

\subsubsection{KG-Based Alignment} 
Suppose in a federated KG each client owns one KG and each KG may have overlapping entities which also exist on other clients. The key technique to improve the performance of KG embedding is to align the embeddings of overlapping entities across KGs \cite{chen2021fede, peng2021fkge, zhang2022fedr}. 
More specifically, after each round $t$, the server collects each local embedding matrix from each client to update the global embedding matrix. Then the server distributes the global embeddings to corresponding clients for subsequent local training. As the first FL framework for KGs, FedE \cite{chen2021fede} enables the server to record all the unique entities from clients with an overall entity table. The server collects the entity embedding matrix $\textbf{E}^t_{(k)}$ of each client $c_k$ and aligns them by 
\begin{equation} \label{fede}
\textbf{E}^t = \left(\mathbb{1} \oslash\sum^M_{k=1}\textbf{v}_k\right) \otimes\sum^M_{k=1}\textbf{P}_{(k)}\textbf{E}^t_{(k)},
\end{equation}
where $\textbf{E}^t$ is the global entity embedding matrix, $\mathbb{1}$ denotes an all-one vector, $\oslash$ denotes element-wise division for vectors and $\otimes$ denotes element-wise multiplication with broadcasting. $\textbf{P}_{(k)}$ denotes client $c_k$'s permutation matrix that maps client $c_k$'s entity matrix to the server's entity table. $\textbf{v}_k$ denotes client $c_k$'s existence vectors.

As the server maintains a complete table of entity embeddings, it can easily infer a relation embedding between two entities $h$ and $t$ by calculating
\begin{equation} \label{infer_relation}
r'=\text{arg} \max\limits_{r} f(h,r,t),
\end{equation}
where $f(\cdot)$ denotes a score function (e.g., TransE \cite{bordes2013transe}) \cite{zhang2022fedr}. To tackle the privacy issue, FedR \cite{zhang2022fedr} was proposed based on relation embedding alignment.

Instead of aligning embeddings on the server, FKGE \cite{peng2021fkge} enables entity alignment between clients. Inspired by PATE-GAN \cite{jordon2018pate}, FKGE involves a privacy-preserving adversarial translation (PPAT) network for adversarial learning. The PPAT network employs a generator as well as a student discriminator and multiple teacher discriminators. The generator first translates aligned entities' embeddings from $\mathcal{G}_k$ into synthesized embeddings and sents them to $\mathcal{G}_s$. The student and teacher discriminators distinguish between the synthesized embeddings and ground truth embeddings in $\mathcal{G}_s$ for each pair of KGs $(\mathcal{G}_k, \mathcal{G}_s)$ which have aligned entities $\mathcal{E}_k \cap \mathcal{E}_s$ and relations $\mathcal{R}_k \cap \mathcal{R}_s$. During the alignment, only synthesized embeddings and gradients are transmitted among clients and data privacy can be guaranteed \cite{guan2021fgnn}.

\subsubsection{User-Item Graph-Based Alignment} 
In a federated recommendation system, each user only has a first-order local user-item subgraph with its own item rating and its neighbors located on its device. A naive method is to align the embeddings of overlapping users and items directly. However, the server can easily infer a user's user-item links by recording the items with non-zero-gradient embeddings from this user because an item embedding gets updated on the user only when the item has the rating score from the user \cite{liu2021fesog}.
%However, it incurs an item embedding gets updated on a user only when the item has the rating score from the user. Therefore, the server can directly infer a user's user-item links by eliminating the items with zero-gradient embeddings from this user, which will lead to privacy leakage \cite{wu2021fedgnn}.

To tackle the privacy leakage, pseudo interacted item sampling and Local Differential Privacy (LDP) \cite{geyer2017ldp} techniques are two common strategies \cite{liu2021fesog, wu2021fedgnn, wu2022fedpergnn}. Before sending gradients to the central server, each user $u_k$ first samples some items that it has not interacted with (i.e., pseudo interacted items). Then it generates embedding gradients of the sampled items (e.g., using a Gaussian distribution) and combines them with the real embedding gradients. Finally, the user applies an LDP module to modify gradients by clipping and adding zero-mean Laplacian noise to gradients 
\begin{equation} \label{ldp}
g\prime_k = \text{clip}(g_k, \delta) + \text{Laplace}(0, \lambda),
\end{equation}
where $g_k$ is the unified gradients of user $u_k$ including model gradients and user/item embedding gradients, $\text{clip}(g_k, \delta)$ denotes limiting $g_k$ with the threshold $\delta$ and $\lambda$ is the strength of Laplacian noise.

\subsection{Non-IID Data Adaptation}
The data distribution on each client may diverge a lot both in node features and graph structures \cite{xie2021gcfl}. Such data heterogeneity may lead to severe model divergence in the federated setting and therefore degrade the performance of the global model. The intuition of mitigating the problem is either to train an effective global model or to train specialized models for each client. The existing techniques handling this problem can be categorized as single global model-based methods and personalized model-based methods.

\subsubsection{Single Global Model-Based Methods}
The goal of single global model-based methods is to train a global graph machine learning model over graph data from clients. The existing techniques tackle non-IID data across clients by designing loss functions and reweighting FL aggregations and interpolating local models.

\noindent \textbf{Loss Function Designing.} The intuition of loss function designing is to replace the original loss function, which is just for high model utility, with a new well-designed loss function that is also targeted at data heterogeneity. A common strategy is to add regularization terms into the local loss function. For instance, to deal with relational data (e.g., KGs) heterogeneity across clients, FedAlign \cite{lin2020fedalign} minimizes the average Optimal Transportation (OT) distance \cite{villani2009ot} between the basis matrices in basis decomposition \cite{schlichtkrull2018rgcn} among clients. Mathematically, to train an $L$-layer graph machine learning model, the regularization term on client $c_k$ can be rewritten as
\begin{equation} \label{fedalign}
    L_k^{r}= \frac{\mu}{M}\sum_{k\neq s}^{M} \sum_{l=1}^{L} \text{OT}(\textbf{V}^l_{(k)}, \textbf{V}^l_{(s)}) + \lambda(||\nabla L_k(\omega)||_2-1)^2,
\end{equation}
where $\textbf{V}^l_{(k)}$ is the basis of $l$-th layer of the local graph machine learning model on client $c_k$ and $\text{OT}(\cdot)$ computes OT distance. The second term is a weight penalty to make the objective function quasi-Lipschitz continuous. $\mu$ and $\lambda$ are hyperparameters to adjust the contributions of each term.

In addition to regularization, another strategy in loss function designing is instance reweighting. For instance, FILT+ \cite{zhu2021filt} pulls the local model closer to the global by minimizing the loss discrepancy between a local model and the global model. Specifically, FILT+ reweights instances on client $c_k$ by putting more weights on samples with less confidence in the loss function
\begin{equation} \label{filt1}
L_k^{u}=\sum_{i=1}^{N_k} (1-\text{exp}(-\Phi(\textbf{x}_i, \omega, \omega_k)))^\gamma l_k(\hat{y}_i, y_i; \omega_k),
\end{equation}
where $\Phi(\cdot)$ is defined as 
\begin{equation} \label{filt2}
\Phi(\textbf{x}_i, \omega, \omega_k)=\phi(\textbf{x}_i, \omega_k) + \text{max}(\phi(\textbf{x}_i, \omega_k)-\phi(\textbf{x}_i, \omega), 0).
\end{equation}
Here $\gamma$ is a hyperparameter and $\phi(\textbf{x}_i, \omega)$ indicates the uncertainty of training sample $\textbf{x}_i$ under the model $\omega$. Generally, if the local model $\omega_k$ on client $c_k$ is less confident about a sample $\textbf{x}_i$ than the global model $\omega$, this sample will obtain a higher weight in the objective function.

Inspired by the model-agnostic meta-learning (MAML) \cite{finn2017maml}, some meta learning-based methods in FGML \cite{li2022fml_st, wang2020graphfl} rewrite the loss function on each client $c_k$ as
\begin{equation} \label{graphfl}
L_k^{u}=\sum_{i=1}^{N_k} l_k(\omega-\alpha \nabla l_k(\omega)),
\end{equation}
where $\alpha$ is a hyperparameter. Although the meta learning-based methods do not minimize the discrepancy between local models and the global model, they find an initial global model which can be easily adapted by clients after performing one or a few extra local updates.

\noindent \textbf{FL Aggregation Reweighting.} Apart from the loss function designing, reweighting local models during FL aggregation is also a solution to deal with Non-IID data. FedGCN \cite{hu2022fedgcn} tries to reweight local model parameters via an attention mechanism. Considering an $L$-layer model $f_\omega$, FedGCN assigns adaptive weights $\{\beta^{t,l}_k\}_{l=1}^L$ to the model parameter  $\omega_k$ from each client $c_k$ in round $t$ for FL aggregation
\begin{equation} \label{fedgcn1}
\omega^{t+1, l} = \sum_{k=1}^M \beta^{t,l}_k\omega^{t,l}_k
\end{equation}
for $l=1,2,\cdots, L$. $\beta^{t,l}_k$ can be calculated through a softmax operation of score function $\alpha^{t,l}_k$
\begin{equation} \label{fedgcn2}
\alpha^{t,l}_k = \text{Attn}(\omega^{t,l}_k,\omega^{t,l})=\textbf{p}^l_k [\omega^{t,l}_k;\omega^{t,l}],
\end{equation}
where $\text{Attn}(\cdot)$ is the attention mechanism, $[\cdot; \cdot]$ indicates a concat operation, and $\textbf{p}^l_k$ is a trainable vector. As a result, $\{\beta^{t,l}_k\}_{l=1}^L$ can dynamically measure the closeness between each local model $\omega_k$ and the global model $\omega$.

\noindent \textbf{Model Interpolation.} The model interpolation technique developed based on parameter weighted average of the global and the local models. Specifically, the model $\omega_k^t$ on client $c_k$ is a combination of its local model $\omega_k^{t-1}$ and the global model $\omega^t$ \cite{zheng2021asfgnn}
\begin{equation}
\omega_k^t = \alpha_k\omega_k^{t-1} + (1-\alpha_k) \omega^t,
\end{equation}
where $\alpha_k$ is a mixing weight. The authors of \cite{zheng2021asfgnn} calculate it as Jensen–Shannon divergence \cite{lin1990js} between local and global data distributions.

\subsubsection{Personalized Model-Based Methods}
Unlike training a single global graph machine learning model, the goal of learning personalized models is to train personalized graph machine learning models for each client. The resulting personalized models are tailored for specific clients and thus result in good performance. Formally, the objective function for training personalized graph machine learning models can be rewritten as 
\begin{equation}  \label{pfl}
    \min\limits_{\omega_1, \omega_2,\cdots, \omega_M}\sum_{k=1}^{M} \frac{N_k}{N} L_k(\omega_k).
\end{equation}

One common strategy for training personalized graph machine learning models is client clustering \cite{ghosh2020ifac, sattler2020cfl, zhang2021ctfed}. The intuition of client clustering is that the clients with similar data distribution can be clustered in a group and the clients in a group share the same model parameters. The basic idea of client clustering in FGML is to dynamically assign clients to multiple clusters based on their latest gradients of graph machine learning models \cite{xie2021gcfl, zhang2022ctfl}. One problem in this idea is that the clustering result is significantly influenced by the latest gradients from clients, which are usually unstable during local training \cite{xie2021gcfl}. GCFL+ solves this problem by taking series of gradient norms into account for client clustering. Unlike collecting parameters in a cluster-wise manner in GCFL, CTFL \cite{zhang2022ctfl} updates the global model based on representative clients of each cluster. A client's local model is updated as the average of the global model and the representative model of the cluster which includes the client.

\section{Structured FL}
\begin{figure}
\centering
\includegraphics[width=\linewidth]{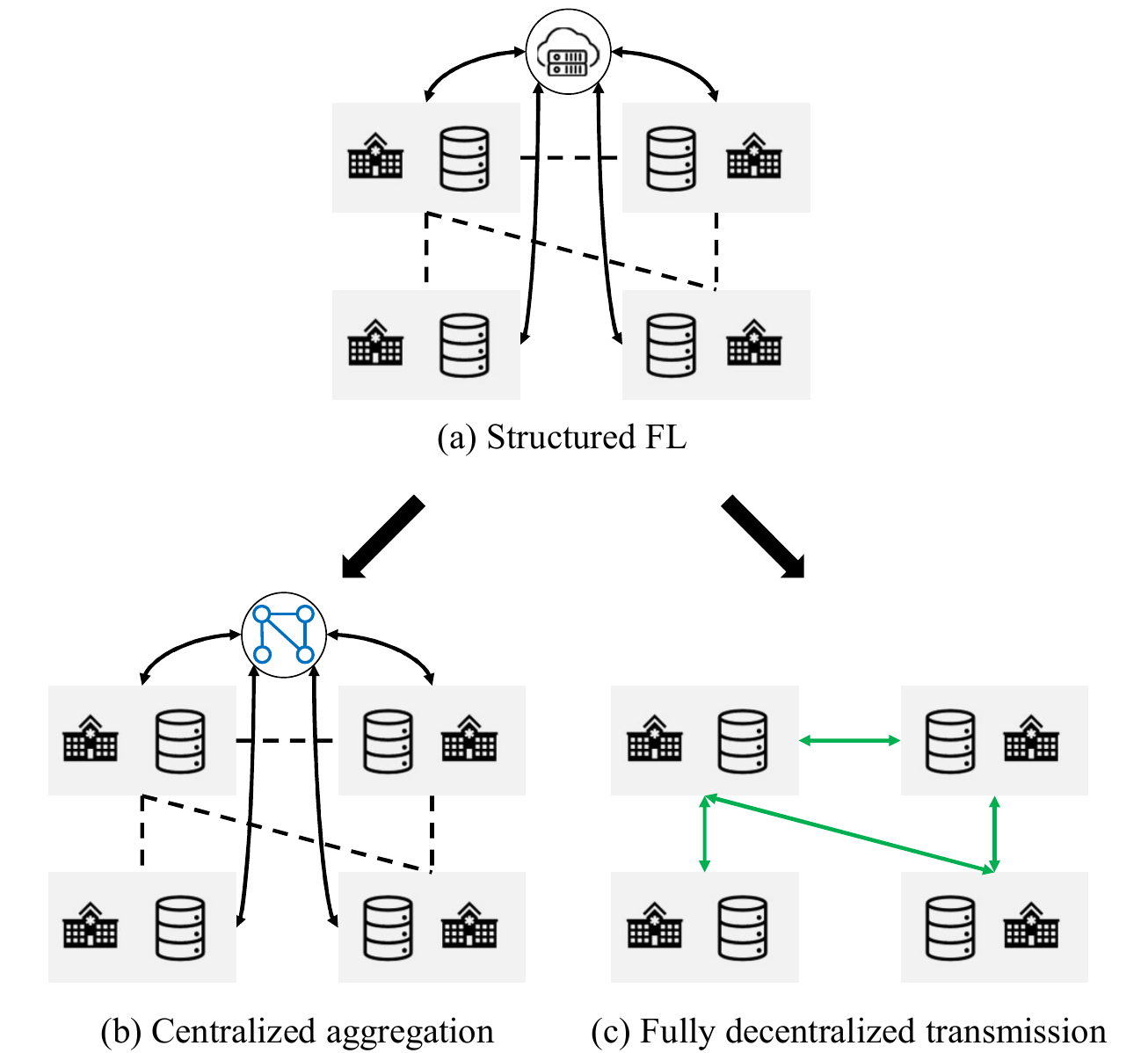}
\caption{The taxonomy of techniques in structured FL. The techniques in (a) structured FL can be categorized as (b) centralized aggregation and (c) fully decentralized transmission. In (b) centralized aggregation, the server performs the FL aggregation and computes global client embeddings based on the client graph. In (c) fully decentralized transmission, parameters are transmitted among clients based on the client graph.}
\label{fig:structure4}
\end{figure}

In the real world, a client may have connections with others, such as road paths existing among traffic sensors. These connections usually contain rich information (e.g., the similarity of data distribution) among clients. Considering these connections, the clients can form a client graph. Structured FL takes the client graph $\mathcal{G}^\mathcal{C}=\{\mathcal{V}^\mathcal{C}, \mathcal{E}^\mathcal{C}\}$ into account and enables a client to obtain information from its neighbors. The key techniques in structured FL can be categorized as centralized aggregation and fully decentralized transmission.  Fig.~\ref{fig:structure4} shows the taxonomy of techniques in structured FL.

\subsection{Centralized Aggregation}
In structured FL with the central server, there exists structural information among clients. It is natural for the server to consider the structural information while updating parameters for each client. Generally, the server first collects parameters from clients as it does in standard FL. Then it updates parameters for each client through a graph machine learning model based on the client graph $\mathcal{G}^\mathcal{C}$ and finally sends the updated parameters back to clients.

\noindent \textbf{Transmitting Model Parameters.} Transmitting local model parameters to the central server as the input of graph machine learning models is a straightforward strategy. For instance, the central server in SFL \cite{chen2022sfl} collects local model parameters $\{\omega_k^t\}_{k=1}^{M}$ from clients in round $t$ and employs a GCN \cite{kipf2016gcn} to compute a graph-based local model parameters $\{\phi_k^{t+1}\}_{k=1}^{M}$ by
\begin{equation}
    \{\phi_k^{t+1}\}_{k=1}^{M} \leftarrow \text{GCN}(\textbf{A}^\mathcal{C}, \{\omega_k^t\}_{k=1}^{M}),
\end{equation}
where $\textbf{A}^\mathcal{C}$ is the adjacency matrix of $\mathcal{G}^\mathcal{C}$ and GCN$(\cdot)$ represents the operations in GCN. The global model parameters $\omega^{t+1}$ are calculated by a readout$(\cdot)$ operation
\begin{equation}
    \omega^{t+1} = \text{readout}(\{\phi_k^{t+1}\}_{k=1}^{M}). 
\end{equation}
Each client $c_k$ updates local mode parameters in round $t$ by minimizing the local loss
\begin{equation}
    \min l_k(\omega_k)+ \lambda [R(\omega_k^t, \omega^t)+R(\omega_k^t, \phi_k^t)],
\end{equation}
where $R(\cdot)$ is a regularization term.

Similarly, the server in BiG-Fed \cite{xingbigfed} collects local model parameters $\{\omega_k^t\}_{k=1}^{M}$ as client representatives and computes a global client embedding $\textbf{H}$ through a graph machine learning model based on $\{\omega_k^t\}_{k=1}^{M}$ and the client graph. Inspired by contrastive learning \cite{chen2020simclr}, Big-Fed optimizes the model by
\begin{equation} \label{bigfed}
\begin{aligned}
        \min \frac{1}{n}&\sum_{k=1}^M\sum_{c_s\in N(c_k)} 1-\cos(\textbf{H}_k, \textbf{H}_s) \\ 
        + \frac{1}{n}&\sum_{k=1}^M \mathbb{E}_{c_s \sim P_k} \max (0, \cos(\textbf{H}_k, \textbf{H}_s)),
\end{aligned}
\end{equation}
where $\textbf{H}_k$ is the global embedding of client $c_k$ and $P_k$ is a client sampling strategy. By minimizing the objective function in Eq.~(\ref{bigfed}), each client will obtain a local model closer to its neighbors.

\noindent \textbf{Transmitting Embeddings.} Instead of gathering model parameters, another strategy for the server is to collect local embeddings from each client and compute global embeddings through graph machine learning models. For instance, CNFGNN \cite{meng2021cnfgnn} assumes that each client $c_k$ represents a sensor with time series data $\textbf{x}_k$. A client $c_k$ first computes a temporal embedding $\textbf{h}_k$ by a local encoder (e.g., GRU \cite{cho2014gru}) which models local temporal dynamics. The central server collects temporal embeddings $\{\textbf{h}_k\}_{k=1}^M$ from clients and employs a graph machine learning model to compute the spatial embeddings $\{\textbf{h}^{\mathcal{G}}_k\}_{k=1}^M$ of clients based on their temporal embeddings and client graph $\mathcal{G}^\mathcal{C}$. As a result, $\textbf{h}^{\mathcal{G}}_k$ integrates information from client $c_k$'s neighbors and involves spatial dynamics. $\textbf{h}^{\mathcal{G}}_k$ is finally sent back to each client $c_k$ as the input of a local decoder with $\textbf{h}_k$ to make prediction.

\subsection{Fully Decentralized Transmission}
In the federated setting, one crucial bottleneck lies in high communication cost on the central server \cite{he2021spreadgnn}. A feasible solution to tackle this issue is to train a model in a fully decentralized fashion. Since the clients in structured FL form a client graph $\mathcal{G}^\mathcal{C}=\{\mathcal{V}^\mathcal{C}, \mathcal{E}^\mathcal{C}\}$, each client $c_k \in \mathcal{V}^\mathcal{C}$ can transmit parameters with its neighbors $N(c_k)$. 
Specifically, when a client $c_k$ receives model parameters $\{\omega_s^t|c_s\in N(c_k)\}$ from its neighbors $N(c_k)$ \cite{lalitha2019p2p, pei2021dfedgnn, rizk2021gfl}, it can perform GNN aggregation to update its local model parameters $\omega_k^{t+1}$ by
\begin{equation} 
    \omega_k^{t+1} = \text{AGG} (\{\omega_s^t|c_s\in N(c_k)\}),
\end{equation}
where $\text{AGG}(\cdot)$ is the aggregation function. An intuitive method \cite{pei2021dfedgnn, rizk2021gfl} following this strategy is to let each client $c_k$ directly sum up the model parameters from its neighboring clients
\begin{equation} 
    \omega_k^{t+1} = \sum_{c_s\in N(c_k)}\textbf{A}^\mathcal{C}_{ks} \cdot\omega_s^t.
\end{equation}
However, this method results in losing local information of each client since the updated model is fully determined by neighboring information. This issue can be mitigated by involving local information during aggregation \cite{brisimi2018health, dinh2021dfedu, he2021spreadgnn, lu2020dsgd2, xing2020dsgd, zehtabi2022efhc}. For example, the authors of \cite{lu2020dsgd2, lu2019dsgd3} directly add local gradients during aggregation. Formally, the operation can be written as
\begin{equation}
    \omega_k^{t+1} = \sum_{c_s\in N(c_k)} \textbf{A}^\mathcal{C}_{ks} \cdot\omega_s^t - \alpha \nabla L_k(\omega_k^t).
\end{equation}
Some methods \cite{xing2020dsgd} choose to retain local model parameters. Formally, these methods can be written as
\begin{equation}
    \omega_k^{t+1} =\textbf{A}^\mathcal{C}_{kk} \cdot\omega_k^t+\sum_{c_s\in N(c_k)} \textbf{A}^\mathcal{C}_{ks} \cdot\omega_s^t - \alpha \nabla L_k(\omega_k^t).
\end{equation}

\section{Applications}
The number of applications of FGML is greatly increasing in various domains such as transportation, computer vision, recommendation systems, and healthcare. In this section, we elaborate some representative applications of FGML.

\subsection{Transportation} 
Traffic prediction plays an important role in urban computing since it benefits reducing traffic congestion and improving transportation efficiency in smart cities \cite{zheng2014urban}. The target of traffic prediction is to predict traffic speed or traffic flow of regions or road segments based on historical data collected by devices deployed in each region or road segment. Traffic data containing spatial-temporal information can be naturally represented as graphs and used as inputs of graph machine learning models for traffic flow prediction. In FL with structured data, a client typically represents an organization. Each client constructs a local graph including devices on the client as nodes and edges are formulated by their physical properties (e.g., Euclidean distances). The clients jointly train a graph machine learning model for traffic flow prediction \cite{lonare2021fedgcn_traffic, zhang2021ctfed, zhang2021fastgnn, zhang2022ctfl}. In structured FL, a client typically represents a device. The clients can form a graph based on their structure information and graph machine learning models are employed to calculate the embeddings of the clients. Due to the privacy issue, each client uploads its temporal embeddings to the central server instead of its raw data \cite{meng2021cnfgnn, yuan2022fedstn}.

Moreover, other applications of FGML in transportation systems, such as location representation \cite{gurukar2021locationtrails, wu2022location2}, routing planing \cite{ye2021fedte, zeng2021route}, and user mobility anomaly detection \cite{tang2021gumard}, are also attracting increasing attention.

\subsection{Computer Vision}
Most existing applications of FGML in computer vision are categorized as FL with structured data. They mainly focus on image classification and object trajectory prediction \cite{caldarola2021fedcg, chakravarty2021cnn_gnn, jiang2022feddy}. The intuition is to construct graphs on clients which incorporate semantic relationships among classes and objects on each client and embed them via graph machine learning models. For image classification, a graph is constructed on each client to represent classes (or domains) and the connections among them; then the clients jointly train a graph machine learning model to learn class embeddings \cite{caldarola2021fedcg, chakravarty2021cnn_gnn}. Typically, the existing techniques for image classification mainly employ CNN-based methods as the backbone model. Apart from the CNN-based backbone model (e.g., ResNet \cite{he2016resnet} and DenseNet \cite{huang2017densenet}), domain and class-specific features and the GNN models are also transmitted for aggregation during federated optimization. For object trajectory prediction, the idea is to construct a series of dynamic graphs from videos. Each graph represents objects and their spatial relationships in a video frame. Besides aggregating information from nearby objects in the current graph, each object also aggregates dynamic information from those in the previous graph by a dynamic GNN. In a federated setting (e.g., a distributed surveillance system), the dynamic GNN is transmitted for collaborative training \cite{jiang2022feddy}.

\subsection{Recommendation Systems}
Since each user in a federated recommendation system has a first-order local user-item graph, the applications in recommendation systems are naturally categorized as FL with structured data. The downstream task in graph-based federated recommendation systems is to produce high-quality item rating for each user. A graph machine learning model learns to predict unobserved item rating to a user by leveraging the embeddings of users and items in a user's local user-item subgraph with its own item rating. Aside from transmitting model parameters, user embeddings and item embeddings are also collected by the server, which leads to information leakage of item rating becoming a primary concern in federated recommendation systems \cite{liu2021fesog, wu2021fedgnn, wu2022fedpergnn}. This privacy issue is mitigated via adding the embeddings of pseudo interacted items and noise to gradients \cite{liu2021fesog, wu2021fedgnn}.

\subsection{Healthcare}
Medical data such as medical images and disease symptoms are very sensitive and private, which results in medical datasets usually existing in isolated hospitals and medical institutes. In general, we take the hospitals and medical institutes as clients and the medical datasets form graphs on these clients; therefore, we categorize related applications in healthcare as FL with structured data.  The applications of FGML in healthcare are targeted at disease and hospitalization prediction based on medical images or health records stored in different hospitals and institutes. One key feature of healthcare applications is complex connections among medical images or health records because of patient interactions. The common strategy of modeling the connections is to construct a graph where each node represents an image or a medical record from a patient \cite{thakur2021mtl}. Due to the privacy issue, the graph is split into multiple subgraphs located in different hospitals and institutes. Several papers deal with cross-hospital links either by reconstructing cross-hospital connections \cite{peng2021fedni} or by transmitting parameters with nearby hospitals \cite{brisimi2018health, lu2020dsgd2, lu2019dsgd3}. Fed-CBT \cite{bayram2021fedcbt} learns connectional brain template representations by modeling them as graphs and fuses the graphs of one subject with a deep graph normalizer (DGN) \cite{gurbuz2020dgn}. STFL \cite{lou2021stfl} takes each channel in polysomnography recordings as a node in a graph and trains a graph machine learning model for federated graph classification.

\subsection{Other Applications}
Apart from the aforementioned applications, FGML has manifold applications in other domains. A number of related papers have explored applications of FL with structured data to various problems, such as human activity recognition \cite{sarkar2021grafehty}, neural architecture search \cite{wang2021agcns}, packet routing \cite{mai2021packet},  malware detection \cite{dam2022malware1, payne2019malware2}, drug discovery \cite{manu2021fl_disco}, and financial crime detection \cite{suzumura2019crime}.

\begin{table*}

\centering
\caption{Summary of graph datasets.}
\begin{tabular}{c|ccccc} \hline
Category & Datasets & \# Graphs & (Avg.) \# Nodes & (Avg.) \# Edges & Works Citation \\ \hline
\multirow{1}{*}{} & CORA \cite{mccallum2000cora} & 1 & 2,708 & 5,429 & \cite{zhang2021fedsage, chen2021fedgraph, liu2021glint, ni2021fedvgcn, zhou2020vfgnn, wang2020graphfl} \\ \cline{2-6}
\multirow{1}{*}{Citation} & CITESEER \cite{giles1998citeseer} & 1 & 4,230 & 5,358 & \cite{zhang2021fedsage, chen2021fedgraph, ni2021fedvgcn, zhou2020vfgnn, wang2020graphfl}\\ \cline{2-6}
\multirow{1}{*}{Networks} & PUBMED \cite{sen2008pubmed} & 1 & 19,717 & 44,338 & \cite{zhang2021fedsage, chen2021fedgraph, liu2021glint, zhang2021ppsgcn, ni2021fedvgcn, zhou2020vfgnn}\\ \cline{2-6}
\multirow{4}{*}{} & arXiv \cite{hu2020ogb} & 1 & 169,343 & 2,315,598 & \cite{zhou2020vfgnn} \\ \cline{2-6}\hline

\multirow{2}{*}{Coauthor} & CS \cite{shchur2018pitfalls} & 1 & 18,333 & 81,894 & \cite{wang2020graphfl, zhang2021fedsage} \\ \cline{2-6} 
\multirow{2}{*}{Networks} & Physics \cite{shchur2018pitfalls} & 1 & 34,493 & 247,962 & \cite{liu2021glint} \\ \cline{2-6} 
\multirow{2}{*}{} & Aminer \cite{tang2008arnetminer} & 1 & 3,923 & 9,023 & \cite{mei2019sgnn} \\ \cline{2-6}\hline 

\multirow{8}{*}{} & KarateClub \cite{zachary1977KarateClub} & 1 & 34 & 156 & \cite{xingbigfed} \\ \cline{2-6} 

\multirow{8}{*}{} & REDDIT \cite{hamilton2017graphsage} & 1 & 232,965 & 114,848,857 & \cite{chen2021fedgraph, liu2021glint, zhang2021ppsgcn} \\ \cline{2-6} 
\multirow{8}{*}{} & REDDIT-BINARY \cite{morris2020tudataset}   & 2,000 & 429.63 & 497.75 & \cite{hu2022fedgcn} \\ \cline{2-6} 
\multirow{1}{*}{Social} & GITHUB \cite{morris2020tudataset} & 12,725 & 113.79 & 234.64 & \cite{hu2022fedgcn} \\ \cline{2-6} 
\multirow{1}{*}{Networks} & IMDB-BINARY \cite{morris2020tudataset}      & 1,000 & 19.77 & 96.53 & \cite{hu2022fedgcn, xie2021gcfl} \\ \cline{2-6} 
\multirow{8}{*}{} & IMDB-MULTI \cite{morris2020tudataset}      & 1,500 & 13 & 65.94 & \cite{hu2022fedgcn, xie2021gcfl} \\ \cline{2-6} 
\multirow{8}{*}{} & COLLAB \cite{morris2020tudataset}          & 5,000 & 74.49 & 2457.78 & \cite{hu2022fedgcn, xie2021gcfl} \\ \cline{2-6} 
\multirow{8}{*}{} & Amazon2M \cite{hu2020ogb} & 1 & 2,449,029 & 61,859,140 & \cite{wang2020graphfl, zhang2021ppsgcn} \\ \cline{2-6} \hline

\multirow{19}{*}{Molecules} & FreeSolv \cite{wu2018moleculenet}    & 642   & 8.72 & 25.6 & \cite{zhu2021filt, pei2021dfedgnn} \\ \cline{2-6}
\multirow{19}{*}{} & Lipophilicity \cite{wu2018moleculenet} & 4,200  & 27.04 & 86.04  & \cite{zhu2021filt, pei2021dfedgnn} \\ \cline{2-6}
\multirow{19}{*}{} & ESOL \cite{wu2018moleculenet}           & 1,128  & 13.29 & 40.65 & \cite{zhu2021filt, pei2021dfedgnn} \\ \cline{2-6}
\multirow{19}{*}{} & MUV \cite{wu2018moleculenet}           & 93,087 & 24.23 & 76.80 & \cite{he2021spreadgnn} \\ \cline{2-6}
\multirow{19}{*}{} & QM8 \cite{wu2018moleculenet}           & 21,786 & 7.77 & 23.95 & \cite{he2021spreadgnn} \\ \cline{2-6}
\multirow{19}{*}{} & QM9 \cite{wu2018moleculenet, morris2020tudataset}           & 133,885 & 8.8 & 27.6 & \cite{zhu2021filt} \\ \cline{2-6}
\multirow{19}{*}{} & Tox21 \cite{wu2018moleculenet, morris2020tudataset}         & 7,831  & 18.51 & 25.94 & \cite{zhu2021filt, pei2021dfedgnn, he2021spreadgnn} \\ \cline{2-6}
\multirow{19}{*}{} & SIDER \cite{wu2018moleculenet}         & 1,427  & 33.64 & 35.36 & \cite{zhu2021filt, pei2021dfedgnn, he2021spreadgnn} \\ \cline{2-6}
\multirow{1}{*}{} & ClinTox \cite{wu2018moleculenet}       & 1,478  & 26.13 & 27.86 & \cite{zhu2021filt, pei2021dfedgnn} \\ \cline{2-6}
\multirow{1}{*}{} & BBBP \cite{wu2018moleculenet}          & 2,039  & 24.05 & 25.94 & \cite{zhu2021filt, pei2021dfedgnn} \\ \cline{2-6}
\multirow{1}{*}{} & BACE \cite{wu2018moleculenet}          & 1,513  & 34.12 & 36.89 & \cite{zhu2021filt, pei2021dfedgnn} \\ \cline{2-6}
\multirow{1}{*}{} & hERG \cite{wu2018moleculenet}           & 10,572 & 29.39 & 94.09 & \cite{pei2021dfedgnn} \\ \cline{2-6}
\multirow{1}{*}{} & NCI1 \cite{morris2020tudataset}           & 4,110  & 29.87 & 32.3 & \cite{hu2022fedgcn, xie2021gcfl} \\ \cline{2-6}
\multirow{1}{*}{} & MUTAG \cite{morris2020tudataset}          & 188 & 17.93 & 19.79 & \cite{xie2021gcfl} \\ \cline{2-6}
\multirow{1}{*}{} & BZR \cite{morris2020tudataset}            & 405 & 35.75 & 38.36 & \cite{xie2021gcfl} \\ \cline{2-6}
\multirow{1}{*}{} & COX2 \cite{morris2020tudataset}            & 467 & 41.22 & 43.45 & \cite{xie2021gcfl} \\ \cline{2-6}
\multirow{1}{*}{} & DHFR \cite{morris2020tudataset}            & 467 & 42.43 & 44.54 & \cite{xie2021gcfl} \\ \cline{2-6}
\multirow{1}{*}{} & PTC\_MR \cite{morris2020tudataset} & 344 & 14.29 & 14.69 & \cite{xie2021gcfl} \\ \cline{2-6}
\multirow{1}{*}{} & AIDS \cite{morris2020tudataset}            & 2,000 & 15.69 & 16.20 & \cite{xie2021gcfl} \\ \cline{2-6} \hline

\multirow{3}{*}{Proteins} & DDI \cite{morris2020tudataset}             & 1,178  & 284.32 & 715.66 & \cite{hu2022fedgcn, xie2021gcfl} \\ \cline{2-6}
\multirow{1}{*}{} & PROTEINS \cite{morris2020tudataset}       & 1,113  & 39.06 & 72.82 & \cite{hu2022fedgcn, xie2021gcfl} \\ \cline{2-6}
\multirow{1}{*}{} & ENZYMES \cite{morris2020tudataset}        & 600   & 32.63 & 62.14 & \cite{hu2022fedgcn, xie2021gcfl} \\ \cline{2-6} \hline

\multirow{1}{*}{} & FB15k-237 \cite{dettmers2018convolutional} & 1 & 14,505 & 212,110 & \cite{chen2021fede, zhang2022fedr} \\ \cline{2-6}
\multirow{1}{*}{Knowledge} & WN18RR \cite{dettmers2018convolutional} & 1 & 40,559 & 71,839 & \cite{zhang2022fedr} \\ \cline{2-6}
\multirow{1}{*}{Graphs} & NELL-995 \cite{wang202pathcon} & 1 & 63,917 & 147,465 & \cite{chen2021fede} \\ \cline{2-6}
\multirow{1}{*}{} & DDB14 \cite{wang202pathcon} & 1 & 9,203 & 44,561 & \cite{zhang2022fedr} \\ \cline{2-6} \hline

\multirow{1}{*}{} & Flixster \cite{jamali2010flixster} & 1 & 6,000 & 26,173 & \cite{wu2021fedgnn, wu2022fedpergnn} \\ \cline{2-6}
\multirow{1}{*}{} & Douban \cite{ma2011douban} & 1 & 6,000 & 136,891 & \cite{wu2021fedgnn, wu2022fedpergnn} \\ \cline{2-6}
\multirow{1}{*}{Recommendation} & Yahoo \cite{dror2012yahoo} & 1 & 6,000 & 5,335 & \cite{wu2021fedgnn, wu2022fedpergnn} \\ \cline{2-6}
\multirow{1}{*}{Systems} & MovieLens \cite{miller2003movielens} & 1 & 82,000 & 10,000,054 & \cite{wu2021fedgnn, wu2022fedpergnn} \\ \cline{2-6}

\multirow{1}{*}{} & Ciao \cite{tang2012mtrust} & 1 & 26,678 & 167,320 & \cite{liu2021fesog} \\ \cline{2-6}
\multirow{1}{*}{} & Epinions \cite{tang2012mtrust} &  1 & 35,079 & 225,579 & \cite{liu2021fesog} \\ \cline{2-6}
\multirow{1}{*}{} & Filmtrust \cite{guo2013filmtrust} & 1 & 3,579 & 35,500 & \cite{liu2021fesog} \\ \cline{2-6} \hline

\multirow{7}{*}{Transportation} & Brazil \cite{ribeiro2017struc2vec} & 1 & 131 & 1,038 & \cite{mei2019sgnn} \\ \cline{2-6}
\multirow{1}{*}{} & Europe \cite{ribeiro2017struc2vec} & 1 & 399 & 5,995 & \cite{mei2019sgnn} \\ \cline{2-6}
\multirow{1}{*}{} & PeMSD4 \cite{pemsd} & 1 & 307 & - & \cite{zhang2022ctfl, chen2022sfl} \\ \cline{2-6}
\multirow{1}{*}{} & PeMSD7 \cite{pemsd} & 1 & 288 & - & \cite{zhang2022ctfl} \\ \cline{2-6}
\multirow{1}{*}{} & PeMSD8 \cite{pemsd}  & 1 & 288 & - & \cite{chen2022sfl} \\ \cline{2-6}
\multirow{1}{*}{} & PEMS-BAY \cite{pemsd}  & 1 & 325 & - & \cite{meng2021cnfgnn, chen2022sfl} \\ \cline{2-6}
\multirow{1}{*}{} & METR-LA \cite{jagadish2014metr_la} & 1 & 207 & - & \cite{meng2021cnfgnn, chen2022sfl} \\ \cline{2-6} \hline

\multirow{2}{*}{Healthcare} & ABIDE \cite{di2014abide} & 1 & 1,029 & - & \cite{peng2021fedni} \\ \cline{2-6}
\multirow{1}{*}{} & ADNI \cite{petersen2010adni} & 1 & 911 & - & \cite{peng2021fedni} \\ \cline{2-6}

\hline\end{tabular}
\label{table:dataset}
\end{table*}

\section{Datasets and Platforms}
In this section, we summarize the existing open graph datasets and platforms used for FGML research. 
\subsection{Datasets}
We organize and introduce examples of real-world datasets in Table \ref{table:dataset}. These datasets are categorized by different application domains, such as citation networks, coauthor networks, social networks, molecules, proteins, KGs, recommendation systems, transportation, and healthcare. For each dataset, we provide basic statistical information including the number of graphs, the (average) number of nodes, the (average) number of edges. We also list the corresponding papers that use these datasets.

\subsection{Platforms}
Although a number of platforms \cite{bonawitz2019tff, caldas2018leaf, ziller2021pysyft} facilitate FL applications in multiple domains such as vision and language, only a few of them provide off-the-shelf supports for graph datasets and graph machine learning models. To the best of our knowledge, FedGraphNN \cite{he2021fedgraphnn} and FederatedScope-GNN (FS-G) \cite{wang2022federatedscope_gnn} are the two platforms supporting tasks in FGML among the existing FL platforms.

\noindent \textbf{FedGraphNN.} FedGraphNN is an FL benchmark system for GNNs in FedML ecosystem \cite{he2020fedml}. It contains 36 graph datasets from 7 domains, such as molecules, proteins, KGs, recommendation systems, citation networks, and social networks. As for graph machine learning models, it supports a series of popular GNN-based models, such as GCN \cite{kipf2016gcn}, GAT \cite{velivckovic2017gat}, GraphSage \cite{hamilton2017graphsage}, and GIN \cite{xu2018gin}, implemented via PyTorch Geometric \cite{fey2019pyg}.

\noindent \textbf{FederatedScope-GNN.}
FS-G is built upon an event-driven FL framework FederatedScope \cite{xie2022federatedscope} and it is compatible with various graph machine learning backends. For the ease of benchmarking approaches in FGML, FS-G incorporates two components specifically designed for FGML: GraphDataZoo and GNNModelZoo. A GraphDataZoo integrates a rich collection of splitting mechanisms for dispersing a given standalone graph dataset into multiple clients. A GNNModelZoo integrates a number of algorithms including vanilla graph machine learning models (e.g., GCN \cite{kipf2016gcn}, GAT \cite{velivckovic2017gat}, and GraphSage \cite{hamilton2017graphsage}) and some recent frameworks in FGML, such as FedSage+ \cite{zhang2021fedsage}, GCFL+ \cite{xie2021gcfl}, and Fedgnn \cite{wu2021fedgnn}. FS-G also supports federated hyper-parameter optimization and model personalization for FGML.

\section{Open Challenges and Future Directions}
In this section, we present some limitations in current studies and provide promising directions for future advances.

\noindent \textbf{Data Heterogeneity of Graph Structures.} Unlike the original FL where data distribution of features and labels are considered, the non-IID characteristic of graph structures is also a key challenge. Although a few papers analyze non-IID graph structures across clients \cite{xie2021gcfl}, few of them address this issue completely. Despite some popular approaches designed for mitigating the non-IID characteristic in standard FL \cite{ghosh2020ifac, shamsian2021pflhn, huang2021fedamp}, the approaches in FGML should take graph structures into account.

\noindent \textbf{Secure Aggregation of Instance Embeddings.} As mentioned in Section 3, the central server may collect instance embeddings from clients for instance alignment in FGML, especially in KGs and user-item graphs. The existing studies mainly apply LDP techniques and pseudo instance sampling to alleviate privacy leakage due to embedding transmission \cite{liu2021fesog, wu2021fedgnn}. However, these algorithms can lead to performance degradation because of introducing noise. Thus, designing an effective yet secure aggregation scheme in FGML is still an open problem.

\noindent \textbf{Communication Reduction Strategies in FGML.}
Communication is a critical bottleneck in the federated setting \cite{mcmahan2017fl}. Large communication overhead makes it difficult to train an FL model. In FGML, we should consider both client-level and node-level relationships between different clients and it usually requires extra communication along these relations \cite{he2021spreadgnn,zhang2021ppsgcn}. Therefore, communication might be a more serious bottleneck in FGML. Further applied research on FGML should better consider the communication reduction strategies such as data compression \cite{lin2017deep} and local updating \cite{li2020fl_survey1}.

\noindent \textbf{Fairness in FGML.}
Fairness is an important topic in FL. Without accessing the sensitive information (e.g., gender and race) in different clients, FL models might show distinct bias against some groups of data \cite{ezzeldin2021fairfed}. Furthermore, we want the model to have similar performance in each client in some FL scenarios \cite{cui2021fcfl,li2021ditto}. They are both requirements of fairness in FL. Considering structural information in FL, FGML provides many non-trivial challenges of fairness, e.g., how the structural information affects different fairness metrics in the federated training process. Novel fairness-aware FGML models are greatly expected. 

\noindent \textbf{Poisoning Attacks and Defenses in FGML.} Recently, a few studies about poisoning attacks and defenses have been proposed \cite{chen2022fraudster, xu2022cba}. Apart from poisoning attacks on data features and model parameters in standard FL, poisoning attacks on graph structures can also affect collaborative training in FGML. Designing efficient attacks on graph structures in FGML and defending against such attacks could be a promising topic in security.

\noindent \textbf{Benchmarks and Platforms.} Compared with abundant benchmarks and platforms in standard FL \cite{bonawitz2019tff, caldas2018leaf, xie2022federatedscope, ziller2021pysyft}, applicable benchmarks and platforms in FGML are still at an infant stage. The current two platforms, FederatedScope-GNN \cite{wang2022federatedscope_gnn} and FedGraphNN \cite{he2021fedgraphnn}, lack either graph datasets or off-the-shelf FGML algorithms. In addition, splitting mechanisms in FGML are significantly different from those in standard FL, especially when a graph is split into multiple subgraphs across clients. Practical distributed graph data from the real world is needed for more practical graph partition.

\section{Conclusions}
A large number of powerful graph machine learning models have achieved remarkable success in different domains. However, graph machine learning in a federated setting still faces a series of new challenges and therefore attracts massive attention from both researchers and practitioners. In this paper, we introduce the concepts of two problem settings in FGML. Then we review the current techniques under each setting in detail and introduce applications of FGML from different domains in the real world. We also summarize open graph datasets and platforms in FGML. In the end, some promising future directions are provided.

\section{Acknowledgements}
This work is supported by the National Science Foundation (IIS-2006844, IIS-2144209, IIS-2223769, CNS-2154962, and BCS-2228534), Commonwealth Cyber Initiative (VV-1Q23-007), the JP Morgan Chase Faculty Research Award, the Cisco Faculty Research Award, the 3 Cavaliers seed grant, and the 4-VA collaborative research grant.

\bibliographystyle{abbrv}
\bibliography{sigkddExp}

\begin{thebibliography}{100}

\bibitem{acar2018he}
A.~Acar, H.~Aksu, A.~S. Uluagac, and M.~Conti.
\newblock A survey on homomorphic encryption schemes: Theory and
  implementation.
\newblock {\em ACM Computing Surveys}, 2018.

\bibitem{bayram2021fedcbt}
H.~C. Bayram and I.~Rekik.
\newblock A federated multigraph integration approach for connectional brain
  template learning.
\newblock In {\em ML-CDS}, 2021.

\bibitem{bonawitz2019tff}
K.~Bonawitz, H.~Eichner, W.~Grieskamp, D.~Huba, A.~Ingerman, V.~Ivanov,
  C.~Kiddon, J.~Kone{\v{c}}n{\`y}, S.~Mazzocchi, B.~McMahan, et~al.
\newblock Towards federated learning at scale: System design.
\newblock In {\em MLSys}, 2019.

\bibitem{bordes2013transe}
A.~Bordes, N.~Usunier, A.~Garcia-Duran, J.~Weston, and O.~Yakhnenko.
\newblock Translating embeddings for modeling multi-relational data.
\newblock In {\em NIPS}, 2013.

\bibitem{brisimi2018health}
T.~S. Brisimi, R.~Chen, T.~Mela, A.~Olshevsky, I.~C. Paschalidis, and W.~Shi.
\newblock Federated learning of predictive models from federated electronic
  health records.
\newblock {\em International Journal of Medical Informatics}, 2018.

\bibitem{cai2021link_pred1}
L.~Cai, J.~Li, J.~Wang, and S.~Ji.
\newblock Line graph neural networks for link prediction.
\newblock {\em IEEE Transactions on Pattern Analysis and Machine Intelligence},
  2021.

\bibitem{caldarola2021fedcg}
D.~Caldarola, M.~Mancini, F.~Galasso, M.~Ciccone, E.~Rodola, and B.~Caputo.
\newblock Cluster-driven graph federated learning over multiple domains.
\newblock In {\em CVPR Workshops}, 2021.

\bibitem{caldas2018leaf}
S.~Caldas, S.~M.~K. Duddu, P.~Wu, T.~Li, J.~Kone{\v{c}}n{\`y}, H.~B. McMahan,
  V.~Smith, and A.~Talwalkar.
\newblock Leaf: A benchmark for federated settings.
\newblock In {\em NeurIPS Workshops}, 2019.

\bibitem{chakravarty2021cnn_gnn}
A.~Chakravarty, A.~Kar, R.~Sethuraman, and D.~Sheet.
\newblock Federated learning for site aware chest radiograph screening.
\newblock In {\em ISBI}, 2021.

\bibitem{chen2021fedgl}
C.~Chen, W.~Hu, Z.~Xu, and Z.~Zheng.
\newblock Fedgl: federated graph learning framework with global
  self-supervision.
\newblock {\em arXiv preprint arXiv:2105.03170}, 2021.

\bibitem{chen2021fedgraph}
F.~Chen, P.~Li, T.~Miyazaki, and C.~Wu.
\newblock Fedgraph: Federated graph learning with intelligent sampling.
\newblock {\em IEEE Transactions on Parallel and Distributed Systems}, 2021.

\bibitem{chen2022sfl}
F.~Chen, G.~Long, Z.~Wu, T.~Zhou, and J.~Jiang.
\newblock Personalized federated learning with graph.
\newblock {\em arXiv preprint arXiv:2203.00829}, 2022.

\bibitem{chen2021rs1}
H.~Chen, L.~Wang, Y.~Lin, C.-C.~M. Yeh, F.~Wang, and H.~Yang.
\newblock Structured graph convolutional networks with stochastic masks for
  recommender systems.
\newblock In {\em SIGIR}, 2021.

\bibitem{chen2022fraudster}
J.~Chen, G.~Huang, H.~Zheng, S.~Yu, W.~Jiang, and C.~Cui.
\newblock Graph-fraudster: Adversarial attacks on graph neural network-based
  vertical federated learning.
\newblock {\em IEEE Transactions on Computational Social Systems}, 2022.

\bibitem{chen2021fede}
M.~Chen, W.~Zhang, Z.~Yuan, Y.~Jia, and H.~Chen.
\newblock Fede: Embedding knowledge graphs in federated setting.
\newblock In {\em IJCKG}, 2021.

\bibitem{chen2020simclr}
T.~Chen, S.~Kornblith, M.~Norouzi, and G.~Hinton.
\newblock A simple framework for contrastive learning of visual
  representations.
\newblock In {\em ICML}, 2020.

\bibitem{cheung2021fedsgc}
T.-H. Cheung, W.~Dai, and S.~Li.
\newblock Fedsgc: Federated simple graph convolution for node classification.
\newblock In {\em IJCAI Workshops}, 2021.

\bibitem{cho2014gru}
K.~Cho, B.~van Merrienboer, C.~Gucehre, D.~Bahdanau, F.~Bougares, H.~Schwenk,
  and Y.~Bengio.
\newblock Learning phrase representations using rnn encoder-decoder for
  statistical machine translation.
\newblock In {\em EMNLP}, 2014.

\bibitem{cui2021fcfl}
S.~Cui, W.~Pan, J.~Liang, C.~Zhang, and F.~Wang.
\newblock Addressing algorithmic disparity and performance inconsistency in
  federated learning.
\newblock In {\em NeurIPS}, 2021.

\bibitem{dam2022malware1}
K.~H.~T. Dam, C.-H.~B. Van~Ouytsel, and A.~Legay.
\newblock Symbolic analysis meets federated learning to enhance malware
  identifier.
\newblock {\em arXiv preprint arXiv:2204.14159}, 2022.

\bibitem{daza2021link_pred2}
D.~Daza, M.~Cochez, and P.~Groth.
\newblock Inductive entity representations from text via link prediction.
\newblock In {\em WWW}, 2021.

\bibitem{dettmers2018convolutional}
T.~Dettmers, P.~Minervini, P.~Stenetorp, and S.~Riedel.
\newblock Convolutional 2d knowledge graph embeddings.
\newblock In {\em AAAI}, 2018.

\bibitem{di2014abide}
A.~Di~Martino, C.-G. Yan, Q.~Li, E.~Denio, F.~X. Castellanos, K.~Alaerts, J.~S.
  Anderson, M.~Assaf, S.~Y. Bookheimer, M.~Dapretto, et~al.
\newblock The autism brain imaging data exchange: Towards a large-scale
  evaluation of the intrinsic brain architecture in autism.
\newblock {\em Molecular psychiatry}, 2014.

\bibitem{dinh2021dfedu}
C.~T. Dinh, T.~T. Vu, N.~H. Tran, M.~N. Dao, and H.~Zhang.
\newblock A new look and convergence rate of federated multi-task learning with
  laplacian regularization.
\newblock {\em arXiv preprint arXiv:2102.07148}, 2021.

\bibitem{dror2012yahoo}
G.~Dror, N.~Koenigstein, Y.~Koren, and M.~Weimer.
\newblock The yahoo! music dataset and kdd-cup’11.
\newblock In {\em KDD Cup 2011}, 2012.

\bibitem{ezzeldin2021fairfed}
Y.~H. Ezzeldin, S.~Yan, C.~He, E.~Ferrara, and S.~Avestimehr.
\newblock Fairfed: Enabling group fairness in federated learning.
\newblock In {\em NeurIPS Workshops}, 2021.

\bibitem{fey2019pyg}
M.~Fey and J.~E. Lenssen.
\newblock Fast graph representation learning with pytorch geometric.
\newblock In {\em ICLR Workshops}, 2019.

\bibitem{finn2017maml}
C.~Finn, P.~Abbeel, and S.~Levine.
\newblock Model-agnostic meta-learning for fast adaptation of deep networks.
\newblock In {\em ICML}, 2017.

\bibitem{gao2022rs2}
C.~Gao, X.~Wang, X.~He, and Y.~Li.
\newblock Graph neural networks for recommender system.
\newblock In {\em WSDM}, 2022.

\bibitem{geyer2017ldp}
R.~C. Geyer, T.~Klein, and M.~Nabi.
\newblock Differentially private federated learning: A client level
  perspective.
\newblock In {\em NIPS Workshops}, 2017.

\bibitem{ghosh2020ifac}
A.~Ghosh, J.~Chung, D.~Yin, and K.~Ramchandran.
\newblock An efficient framework for clustered federated learning.
\newblock In {\em NeurIPS}, 2020.

\bibitem{giles1998citeseer}
C.~L. Giles, K.~D. Bollacker, and S.~Lawrence.
\newblock Citeseer: An automatic citation indexing system.
\newblock In {\em DL}, 1998.

\bibitem{guan2021fgnn}
Z.~Guan, Y.~Li, Z.~Xue, Y.~Liu, H.~Gao, and Y.~Shao.
\newblock Federated graph neural network for cross-graph node classification.
\newblock In {\em CCIS}, 2021.

\bibitem{guo2013filmtrust}
G.~Guo, J.~Zhang, and N.~Yorke-Smith.
\newblock A novel bayesian similarity measure for recommender systems.
\newblock In {\em IJCAI}, 2013.

\bibitem{gurbuz2020dgn}
M.~B. Gurbuz and I.~Rekik.
\newblock Deep graph normalizer: A geometric deep learning approach for
  estimating connectional brain templates.
\newblock In {\em MICCAI}, 2020.

\bibitem{gurukar2021locationtrails}
S.~Gurukar, S.~Parthasarathy, R.~Ramnath, C.~Calder, and S.~Moosavi.
\newblock Locationtrails: A federated approach to learning location embeddings.
\newblock In {\em ASONAM}, 2021.

\bibitem{hamilton2017graphsage}
W.~Hamilton, Z.~Ying, and J.~Leskovec.
\newblock Inductive representation learning on large graphs.
\newblock In {\em NIPS}, 2017.

\bibitem{hang2021node_class2}
M.~Hang, J.~Neville, and B.~Ribeiro.
\newblock A collective learning framework to boost gnn expressiveness for node
  classification.
\newblock In {\em ICML}, 2021.

\bibitem{he2021fedgraphnn}
C.~He, K.~Balasubramanian, E.~Ceyani, C.~Yang, H.~Xie, L.~Sun, L.~He, L.~Yang,
  P.~S. Yu, Y.~Rong, et~al.
\newblock Fedgraphnn: A federated learning system and benchmark for graph
  neural networks.
\newblock In {\em ICLR Workshops}, 2021.

\bibitem{he2021spreadgnn}
C.~He, E.~Ceyani, K.~Balasubramanian, M.~Annavaram, and S.~Avestimehr.
\newblock Spreadgnn: Serverless multi-task federated learning for graph neural
  networks.
\newblock In {\em AAAI}, 2022.

\bibitem{he2020fedml}
C.~He, S.~Li, J.~So, X.~Zeng, M.~Zhang, H.~Wang, X.~Wang, P.~Vepakomma,
  A.~Singh, H.~Qiu, et~al.
\newblock Fedml: A research library and benchmark for federated machine
  learning.
\newblock In {\em NeurIPS}, 2020.

\bibitem{he2016resnet}
K.~He, X.~Zhang, S.~Ren, and J.~Sun.
\newblock Deep residual learning for image recognition.
\newblock In {\em CVPR}, 2016.

\bibitem{hu2022fedgcn}
K.~Hu, J.~Wu, Y.~Li, M.~Lu, L.~Weng, and M.~Xia.
\newblock Fedgcn: Federated learning-based graph convolutional networks for
  non-euclidean spatial data.
\newblock {\em Mathematics}, 2022.

\bibitem{hu2020ogb}
W.~Hu, M.~Fey, M.~Zitnik, Y.~Dong, H.~Ren, B.~Liu, M.~Catasta, and J.~Leskovec.
\newblock Open graph benchmark: Datasets for machine learning on graphs.
\newblock In {\em NeurIPS}, 2020.

\bibitem{huang2017densenet}
G.~Huang, Z.~Liu, L.~Van Der~Maaten, and K.~Q. Weinberger.
\newblock Densely connected convolutional networks.
\newblock In {\em CVPR}, 2017.

\bibitem{huang2021fedamp}
Y.~Huang, L.~Chu, Z.~Zhou, L.~Wang, J.~Liu, J.~Pei, and Y.~Zhang.
\newblock Personalized cross-silo federated learning on non-iid data.
\newblock In {\em AAAI}, 2021.

\bibitem{jagadish2014metr_la}
H.~V. Jagadish, J.~Gehrke, A.~Labrinidis, Y.~Papakonstantinou, J.~M. Patel,
  R.~Ramakrishnan, and C.~Shahabi.
\newblock Big data and its technical challenges.
\newblock {\em Communications of the ACM}, 2014.

\bibitem{jamali2010flixster}
M.~Jamali and M.~Ester.
\newblock A matrix factorization technique with trust propagation for
  recommendation in social networks.
\newblock In {\em RecSys}, 2010.

\bibitem{jiang2022feddy}
M.~Jiang, T.~Jung, R.~Karl, and T.~Zhao.
\newblock Federated dynamic graph neural networks with secure aggregation for
  video-based distributed surveillance.
\newblock {\em ACM Transactions on Intelligent Systems and Technology}, 2022.

\bibitem{jin2020structure}
W.~Jin, Y.~Ma, X.~Liu, X.~Tang, S.~Wang, and J.~Tang.
\newblock Graph structure learning for robust graph neural networks.
\newblock In {\em KDD}, 2020.

\bibitem{johnson2020structure}
D.~Johnson, H.~Larochelle, and D.~Tarlow.
\newblock Learning graph structure with a finite-state automaton layer.
\newblock In {\em NeurIPS}, 2020.

\bibitem{jordon2018pate}
J.~Jordon, J.~Yoon, and M.~Van Der~Schaar.
\newblock Pate-gan: Generating synthetic data with differential privacy
  guarantees.
\newblock In {\em ICLR}, 2018.

\bibitem{kairouz2021fl_survey2}
P.~Kairouz, H.~B. McMahan, B.~Avent, A.~Bellet, M.~Bennis, A.~N. Bhagoji,
  K.~Bonawitz, Z.~Charles, G.~Cormode, R.~Cummings, et~al.
\newblock Advances and open problems in federated learning.
\newblock {\em Foundations and Trends{\textregistered} in Machine Learning},
  2021.

\bibitem{kipf2016gcn}
T.~N. Kipf and M.~Welling.
\newblock Semi-supervised classification with graph convolutional networks.
\newblock In {\em ICLR}, 2017.

\bibitem{lalitha2019p2p}
A.~Lalitha, O.~C. Kilinc, T.~Javidi, and F.~Koushanfar.
\newblock Peer-to-peer federated learning on graphs.
\newblock {\em arXiv preprint arXiv:1901.11173}, 2019.

\bibitem{li2021traffic1}
M.~Li and Z.~Zhu.
\newblock Spatial-temporal fusion graph neural networks for traffic flow
  forecasting.
\newblock In {\em AAAI}, 2021.

\bibitem{li2021ditto}
T.~Li, S.~Hu, A.~Beirami, and V.~Smith.
\newblock Ditto: Fair and robust federated learning through personalization.
\newblock In {\em ICML}, 2021.

\bibitem{li2020fl_survey1}
T.~Li, A.~K. Sahu, A.~Talwalkar, and V.~Smith.
\newblock Federated learning: Challenges, methods, and future directions.
\newblock {\em IEEE Signal Processing Magazine}, 2020.

\bibitem{li2022fml_st}
W.~Li and S.~Wang.
\newblock Federated meta-learning for spatial-temporal prediction.
\newblock {\em Neural Computing and Applications}, 2022.

\bibitem{lin1990js}
J.~Lin and S.~Wong.
\newblock A new directed divergence measure and its characterization.
\newblock {\em International Journal of General System}, 1990.

\bibitem{lin2020fedalign}
Y.~Lin, C.~Chen, C.~Chen, and L.~Wang.
\newblock Improving federated relational data modeling via basis alignment and
  weight penalty.
\newblock {\em arXiv preprint arXiv:2011.11369}, 2020.

\bibitem{lin2017deep}
Y.~Lin, S.~Han, H.~Mao, Y.~Wang, and W.~J. Dally.
\newblock Deep gradient compression: Reducing the communication bandwidth for
  distributed training.
\newblock In {\em ICLR}, 2018.

\bibitem{liu2022review2}
R.~Liu and H.~Yu.
\newblock Federated graph neural networks: Overview, techniques and challenges.
\newblock {\em arXiv preprint arXiv:2202.07256}, 2022.

\bibitem{liu2021glint}
T.~Liu, P.~Li, and Y.~Gu.
\newblock Glint: Decentralized federated graph learning with traffic throttling
  and flow scheduling.
\newblock In {\em IWQOS}, 2021.

\bibitem{liu2022structure}
Y.~Liu, Y.~Zheng, D.~Zhang, H.~Chen, H.~Peng, and S.~Pan.
\newblock Towards unsupervised deep graph structure learning.
\newblock In {\em WWW}, 2022.

\bibitem{liu2021fesog}
Z.~Liu, L.~Yang, Z.~Fan, H.~Peng, and P.~S. Yu.
\newblock Federated social recommendation with graph neural network.
\newblock {\em ACM Transactions on Intelligent Systems and Technology}, 2021.

\bibitem{lonare2021fedgcn_traffic}
S.~Lonare and R.~Bhramaramba.
\newblock Federated approach for privacy-preserving traffic prediction using
  graph convolutional network.
\newblock {\em Journal of Shanghai Jiaotong University (Science)}, 2021.

\bibitem{lou2021stfl}
G.~Lou, Y.~Liu, T.~Zhang, and X.~Zheng.
\newblock Stfl: A temporal-spatial federated learning framework for graph
  neural networks.
\newblock In {\em AAAI Workshops}, 2021.

\bibitem{lu2020dsgd2}
S.~Lu, Y.~Zhang, and Y.~Wang.
\newblock Decentralized federated learning for electronic health records.
\newblock In {\em CISS}, 2020.

\bibitem{lu2019dsgd3}
S.~Lu, Y.~Zhang, Y.~Wang, and C.~Mack.
\newblock Learn electronic health records by fully decentralized federated
  learning.
\newblock In {\em NeurIPS Workshops}, 2019.

\bibitem{ma2011douban}
H.~Ma, D.~Zhou, C.~Liu, M.~R. Lyu, and I.~King.
\newblock Recommender systems with social regularization.
\newblock In {\em WSDM}, 2011.

\bibitem{mahmood2021bioinformatics1}
O.~Mahmood, E.~Mansimov, R.~Bonneau, and K.~Cho.
\newblock Masked graph modeling for molecule generation.
\newblock {\em Nature communications}, 2021.

\bibitem{mai2021packet}
X.~Mai, Q.~Fu, and Y.~Chen.
\newblock Packet routing with graph attention multi-agent reinforcement
  learning.
\newblock In {\em GLOBECOM}, 2021.

\bibitem{manu2021fl_disco}
D.~Manu, Y.~Sheng, J.~Yang, J.~Deng, T.~Geng, A.~Li, C.~Ding, W.~Jiang, and
  L.~Yang.
\newblock Fl-disco: Federated generative adversarial network for graph-based
  molecule drug discovery.
\newblock In {\em ICCAD}, 2021.

\bibitem{mccallum2000cora}
A.~K. McCallum, K.~Nigam, J.~Rennie, and K.~Seymore.
\newblock Automating the construction of internet portals with machine
  learning.
\newblock {\em Information Retrieval}, 2000.

\bibitem{mcmahan2017fl}
B.~McMahan, E.~Moore, D.~Ramage, S.~Hampson, and B.~A. y~Arcas.
\newblock Communication-efficient learning of deep networks from decentralized
  data.
\newblock In {\em AISTATS}, 2017.

\bibitem{mei2019sgnn}
G.~Mei, Z.~Guo, S.~Liu, and L.~Pan.
\newblock Sgnn: A graph neural network based federated learning approach by
  hiding structure.
\newblock In {\em BigData}, 2019.

\bibitem{meng2021cnfgnn}
C.~Meng, S.~Rambhatla, and Y.~Liu.
\newblock Cross-node federated graph neural network for spatio-temporal data
  modeling.
\newblock In {\em KDD}, 2021.

\bibitem{miller2003movielens}
B.~N. Miller, I.~Albert, S.~K. Lam, J.~A. Konstan, and J.~Riedl.
\newblock Movielens unplugged: Experiences with an occasionally connected
  recommender system.
\newblock In {\em ACM IUI}, 2003.

\bibitem{morris2020tudataset}
C.~Morris, N.~M. Kriege, F.~Bause, K.~Kersting, P.~Mutzel, and M.~Neumann.
\newblock Tudataset: A collection of benchmark datasets for learning with
  graphs.
\newblock In {\em ICML Workshops}, 2020.

\bibitem{ni2021fedvgcn}
X.~Ni, X.~Xu, L.~Lyu, C.~Meng, and W.~Wang.
\newblock A vertical federated learning framework for graph convolutional
  network.
\newblock {\em arXiv preprint arXiv:2106.11593}, 2021.

\bibitem{pemsd}
C.~D. of~Transportation.
\newblock California department of transportation.
\newblock {https://pems.dot.ca.gov/}.

\bibitem{olsen2018dtw}
N.~L. Olsen, B.~Markussen, and L.~L. Raket.
\newblock Simultaneous inference for misaligned multivariate functional data.
\newblock {\em Journal of the Royal Statistical Society: Series C (Applied
  Statistics)}, 2018.

\bibitem{panagopoulos2021healthcare2}
G.~Panagopoulos, G.~Nikolentzos, and M.~Vazirgiannis.
\newblock Transfer graph neural networks for pandemic forecasting.
\newblock In {\em AAAI}, 2021.

\bibitem{payne2019malware2}
J.~Payne and A.~Kundu.
\newblock Towards deep federated defenses against malware in cloud ecosystems.
\newblock In {\em TPS-ISA}, 2019.

\bibitem{pei2021dfedgnn}
Y.~Pei, R.~Mao, Y.~Liu, C.~Chen, S.~Xu, F.~Qiang, and B.~E. Tech.
\newblock Decentralized federated graph neural networks.
\newblock In {\em IJCAI Workshops}, 2021.

\bibitem{peng2021fkge}
H.~Peng, H.~Li, Y.~Song, V.~Zheng, and J.~Li.
\newblock Differentially private federated knowledge graphs embedding.
\newblock In {\em CIKM}, 2021.

\bibitem{peng2021fedni}
L.~Peng, N.~Wang, N.~Dvornek, X.~Zhu, and X.~Li.
\newblock Fedni: Federated graph learning with network inpainting for
  population-based disease prediction.
\newblock {\em IEEE Transactions on Medical Imaging}, 2022.

\bibitem{petersen2010adni}
R.~C. Petersen, P.~Aisen, L.~A. Beckett, M.~Donohue, A.~Gamst, D.~J. Harvey,
  C.~Jack, W.~Jagust, L.~Shaw, A.~Toga, et~al.
\newblock Alzheimer's disease neuroimaging initiative (adni): clinical
  characterization.
\newblock {\em Neurology}, 2010.

\bibitem{ribeiro2017struc2vec}
L.~F. Ribeiro, P.~H. Saverese, and D.~R. Figueiredo.
\newblock struc2vec: Learning node representations from structural identity.
\newblock In {\em KDD}, 2017.

\bibitem{rizk2021gfl}
E.~Rizk and A.~H. Sayed.
\newblock A graph federated architecture with privacy preserving learning.
\newblock In {\em SPAWC}, 2021.

\bibitem{ru2021cdp}
S.~Ru, B.~Zhang, Y.~Jie, C.~Zhang, L.~Wei, and C.~Gu.
\newblock Graph neural networks for privacy-preserving recommendation with
  secure hardware.
\newblock In {\em NaNA}, 2021.

\bibitem{sarkar2021grafehty}
A.~Sarkar, T.~Sen, and A.~K. Roy.
\newblock Grafehty: Graph neural network using federated learning for human
  activity recognition.
\newblock In {\em ICMLA}, 2021.

\bibitem{sattler2020cfl}
F.~Sattler, K.-R. M{\"u}ller, and W.~Samek.
\newblock Clustered federated learning: Model-agnostic distributed multitask
  optimization under privacy constraints.
\newblock {\em IEEE Transactions on Neural Networks and Learning Systems},
  2020.

\bibitem{schlichtkrull2018rgcn}
M.~Schlichtkrull, T.~N. Kipf, P.~Bloem, R.~v.~d. Berg, I.~Titov, and
  M.~Welling.
\newblock Modeling relational data with graph convolutional networks.
\newblock In {\em ESWC}, 2018.

\bibitem{sen2008pubmed}
P.~Sen, G.~Namata, M.~Bilgic, L.~Getoor, B.~Galligher, and T.~Eliassi-Rad.
\newblock Collective classification in network data.
\newblock {\em AI magazine}, 2008.

\bibitem{shamsian2021pflhn}
A.~Shamsian, A.~Navon, E.~Fetaya, and G.~Chechik.
\newblock Personalized federated learning using hypernetworks.
\newblock In {\em ICML}, 2021.

\bibitem{shchur2018pitfalls}
O.~Shchur, M.~Mumme, A.~Bojchevski, and S.~Gunnemann.
\newblock Pitfalls of graph neural network evaluation.
\newblock {\em arXiv preprint arXiv:1811.05868}, 2018.

\bibitem{suzumura2019crime}
T.~Suzumura, Y.~Zhou, N.~Baracaldo, G.~Ye, K.~Houck, R.~Kawahara, A.~Anwar,
  L.~L. Stavarache, Y.~Watanabe, P.~Loyola, et~al.
\newblock Towards federated graph learning for collaborative financial crimes
  detection.
\newblock In {\em NeurIPS Workshops}, 2019.

\bibitem{tan2022fl_survey5}
A.~Z. Tan, H.~Yu, L.~Cui, and Q.~Yang.
\newblock Towards personalized federated learning.
\newblock {\em IEEE Transactions on Neural Networks and Learning Systems},
  2022.

\bibitem{tang2012mtrust}
J.~Tang, H.~Gao, and H.~Liu.
\newblock mtrust: Discerning multi-faceted trust in a connected world.
\newblock In {\em WSDM}, 2012.

\bibitem{tang2008arnetminer}
J.~Tang, J.~Zhang, L.~Yao, J.~Li, L.~Zhang, and Z.~Su.
\newblock Arnetminer: Extraction and mining of academic social networks.
\newblock In {\em KDD}, 2008.

\bibitem{tang2021gumard}
Y.~Tang, R.~Jia, X.~Zhou, Z.~Li, H.~Jin, and C.~Zhao.
\newblock Federated learning of user mobility anomaly based on graph attention
  networks.
\newblock In {\em ICCC}, 2021.

\bibitem{thakur2021mtl}
A.~Thakur, P.~Sharma, and D.~A. Clifton.
\newblock Dynamic neural graphs based federated reptile for semi-supervised
  multi-tasking in healthcare applications.
\newblock {\em IEEE Journal of Biomedical and Health Informatics}, 2021.

\bibitem{velivckovic2017gat}
P.~Veli{\v{c}}kovi{\'c}, G.~Cucurull, A.~Casanova, A.~Romero, P.~Lio, and
  Y.~Bengio.
\newblock Graph attention networks.
\newblock In {\em ICLR}, 2018.

\bibitem{villani2009ot}
C.~Villani.
\newblock {\em Optimal transport: old and new}.
\newblock 2009.

\bibitem{voigt2017gdpr}
P.~Voigt and A.~Von~dem Bussche.
\newblock The eu general data protection regulation (gdpr).
\newblock {\em A Practical Guide, 1st Ed., Cham: Springer International
  Publishing}, 2017.

\bibitem{wang2020graphfl}
B.~Wang, A.~Li, H.~Li, and Y.~Chen.
\newblock Graphfl: A federated learning framework for semi-supervised node
  classification on graphs.
\newblock {\em arXiv preprint arXiv:2012.04187}, 2020.

\bibitem{wang2021agcns}
C.~Wang, B.~Chen, G.~Li, and H.~Wang.
\newblock Fl-agcns: Federated learning framework for automatic graph
  convolutional network search.
\newblock {\em arXiv preprint arXiv:2104.04141}, 2021.

\bibitem{wang202pathcon}
H.~Wang, H.~Ren, and J.~Leskovec.
\newblock Relational message passing for knowledge graph completion.
\newblock In {\em KDD}, 2021.

\bibitem{wang2022federatedscope_gnn}
Z.~Wang, W.~Kuang, Y.~Xie, L.~Yao, Y.~Li, B.~Ding, and J.~Zhou.
\newblock Federatedscope-gnn: Towards a unified, comprehensive and efficient
  package for federated graph learning.
\newblock In {\em KDD}, 2022.

\bibitem{wang202healthcare1}
Z.~Wang, R.~Wen, X.~Chen, S.~Cao, S.-L. Huang, B.~Qian, and Y.~Zheng.
\newblock Online disease diagnosis with inductive heterogeneous graph
  convolutional networks.
\newblock In {\em WWW}, 2021.

\bibitem{wu2021fedgnn}
C.~Wu, F.~Wu, Y.~Cao, Y.~Huang, and X.~Xie.
\newblock Fedgnn: Federated graph neural network for privacy-preserving
  recommendation.
\newblock In {\em ICML Workshops}, 2021.

\bibitem{wu2022fedpergnn}
C.~Wu, F.~Wu, L.~Lyu, T.~Qi, Y.~Huang, and X.~Xie.
\newblock A federated graph neural network framework for privacy-preserving
  personalization.
\newblock {\em Nature Communications}, 2022.

\bibitem{wu2018moleculenet}
Z.~Wu, B.~Ramsundar, E.~N. Feinberg, J.~Gomes, C.~Geniesse, A.~S. Pappu,
  K.~Leswing, and V.~Pande.
\newblock Moleculenet: a benchmark for molecular machine learning.
\newblock {\em Chemical science}, 2018.

\bibitem{wu2022location2}
Z.~Wu, X.~Wu, and Y.~Long.
\newblock Multi-level federated graph learning and self-attention based
  personalized wi-fi indoor fingerprint localization.
\newblock {\em IEEE Communications Letters}, 2022.

\bibitem{xie2021gcfl}
H.~Xie, J.~Ma, L.~Xiong, and C.~Yang.
\newblock Federated graph classification over non-iid graphs.
\newblock In {\em NeurIPS}, 2021.

\bibitem{xie2022federatedscope}
Y.~Xie, Z.~Wang, D.~Chen, D.~Gao, L.~Yao, W.~Kuang, Y.~Li, B.~Ding, and
  J.~Zhou.
\newblock Federatedscope: A comprehensive and flexible federated learning
  platform via message passing.
\newblock {\em arXiv preprint arXiv:2204.05011}, 2022.

\bibitem{xing2020dsgd}
H.~Xing, O.~Simeone, and S.~Bi.
\newblock Decentralized federated learning via sgd over wireless d2d networks.
\newblock In {\em SPAWC}, 2020.

\bibitem{xingbigfed}
P.~Xing, S.~Lu, L.~Wu, and H.~Yu.
\newblock Big-fed: Bilevel optimization enhanced graph-aided federated
  learning.
\newblock In {\em ICML Workshops}, 2021.

\bibitem{xu2022cba}
J.~Xu, R.~Wang, K.~Liang, and S.~Picek.
\newblock More is better (mostly): On the backdoor attacks in federated graph
  neural networks.
\newblock {\em arXiv preprint arXiv:2202.03195}, 2022.

\bibitem{xu2018gin}
K.~Xu, W.~Hu, J.~Leskovec, and S.~Jegelka.
\newblock How powerful are graph neural networks?
\newblock In {\em ICLR}, 2019.

\bibitem{yang2019fl_survey3}
Q.~Yang, Y.~Liu, T.~Chen, and Y.~Tong.
\newblock Federated machine learning: Concept and applications.
\newblock {\em ACM Transactions on Intelligent Systems and Technology}, 2019.

\bibitem{ye2021fedte}
M.~Ye, J.~Zhang, Z.~Guo, and H.~J. Chao.
\newblock Federated traffic engineering with supervised learning in
  multi-region networks.
\newblock In {\em ICNP}, 2021.

\bibitem{yuan2022fedstn}
X.~Yuan, J.~Chen, J.~Yang, N.~Zhang, T.~Yang, T.~Han, and A.~Taherkordi.
\newblock Fedstn: Graph representation driven federated learning for edge
  computing enabled urban traffic flow prediction.
\newblock {\em IEEE Transactions on Intelligent Transportation Systems}, 2022.

\bibitem{zachary1977KarateClub}
W.~W. Zachary.
\newblock An information flow model for conflict and fission in small groups.
\newblock {\em Journal of anthropological research}, 1977.

\bibitem{zehtabi2022efhc}
S.~Zehtabi, S.~Hosseinalipour, and C.~G. Brinton.
\newblock Decentralized event-triggered federated learning with heterogeneous
  communication thresholds.
\newblock {\em arXiv preprint arXiv:2204.03726}, 2022.

\bibitem{zeng2021route}
T.~Zeng, J.~Guo, K.~J. Kim, K.~Parsons, P.~Orlik, S.~Di~Cairano, and W.~Saad.
\newblock Multi-task federated learning for traffic prediction and its
  application to route planning.
\newblock In {\em IV}, 2021.

\bibitem{zhang2021ppsgcn}
B.~Zhang, M.~Luo, S.~Feng, Z.~Liu, J.~Zhou, and Q.~Zheng.
\newblock Ppsgcn: A privacy-preserving subgraph sampling based distributed gcn
  training method.
\newblock {\em arXiv preprint arXiv:2110.12906}, 2021.

\bibitem{zhang2021ctfed}
C.~Zhang, L.~Cui, S.~Yu, and J.~James.
\newblock A communication-efficient federated learning scheme for iot-based
  traffic forecasting.
\newblock {\em IEEE Internet of Things Journal}, 2021.

\bibitem{zhang2021fastgnn}
C.~Zhang, S.~Zhang, J.~James, and S.~Yu.
\newblock Fastgnn: A topological information protected federated learning
  approach for traffic speed forecasting.
\newblock {\em IEEE Transactions on Industrial Informatics}, 2021.

\bibitem{zhang2022ctfl}
C.~Zhang, S.~Zhang, S.~Yu, and J.~James.
\newblock Graph-based traffic forecasting via communication-efficient federated
  learning.
\newblock In {\em WCNC}, 2022.

\bibitem{zhang2021review1}
H.~Zhang, T.~Shen, F.~Wu, M.~Yin, H.~Yang, and C.~Wu.
\newblock Federated graph learning--a position paper.
\newblock {\em arXiv preprint arXiv:2105.11099}, 2021.

\bibitem{zhang2022fedr}
K.~Zhang, Y.~Wang, H.~Wang, L.~Huang, C.~Yang, and L.~Sun.
\newblock Efficient federated learning on knowledge graphs via
  privacy-preserving relation embedding aggregation.
\newblock In {\em ACL Workshops}, 2022.

\bibitem{zhang2021fedsage}
K.~Zhang, C.~Yang, X.~Li, L.~Sun, and S.~M. Yiu.
\newblock Subgraph federated learning with missing neighbor generation.
\newblock In {\em NeurIPS}, 2021.

\bibitem{zhang2021traffic2}
X.~Zhang, C.~Huang, Y.~Xu, L.~Xia, P.~Dai, L.~Bo, J.~Zhang, and Y.~Zheng.
\newblock Traffic flow forecasting with spatial-temporal graph diffusion
  network.
\newblock In {\em AAAI}, 2021.

\bibitem{zhang2021bioinformatics2}
Z.~Zhang, Q.~Liu, H.~Wang, C.~Lu, and C.-K. Lee.
\newblock Motif-based graph self-supervised learning for molecular property
  prediction.
\newblock In {\em NeurIPS}, 2021.

\bibitem{zhao2021data_aug}
T.~Zhao, Y.~Liu, L.~Neves, O.~Woodford, M.~Jiang, and N.~Shah.
\newblock Data augmentation for graph neural networks.
\newblock In {\em AAAI}, 2021.

\bibitem{zhao2021node_class1}
T.~Zhao, X.~Zhang, and S.~Wang.
\newblock Graphsmote: Imbalanced node classification on graphs with graph
  neural networks.
\newblock In {\em WSDM}, 2021.

\bibitem{zheng2021asfgnn}
L.~Zheng, J.~Zhou, C.~Chen, B.~Wu, L.~Wang, and B.~Zhang.
\newblock Asfgnn: Automated separated-federated graph neural network.
\newblock {\em Peer-to-Peer Networking and Applications}, 2021.

\bibitem{zheng2014urban}
Y.~Zheng, L.~Capra, O.~Wolfson, and H.~Yang.
\newblock Urban computing: Concepts, methodologies, and applications.
\newblock {\em ACM Transactions on Intelligent Systems and Technology}, 2014.

\bibitem{zhou2020vfgnn}
J.~Zhou, C.~Chen, L.~Zheng, H.~Wu, J.~Wu, X.~Zheng, B.~Wu, Z.~Liu, and L.~Wang.
\newblock Vertically federated graph neural network for privacy-preserving node
  classification.
\newblock In {\em IJCAI}, 2022.

\bibitem{zhu2021fl_survey4}
H.~Zhu, J.~Xu, S.~Liu, and Y.~Jin.
\newblock Federated learning on non-iid data: A survey.
\newblock {\em Neurocomputing}, 2021.

\bibitem{zhu2021filt}
W.~Zhu, A.~White, and J.~Luo.
\newblock Federated learning of molecular properties in a heterogeneous
  setting.
\newblock {\em arXiv preprint arXiv:2109.07258}, 2021.

\bibitem{ziller2021pysyft}
A.~Ziller, A.~Trask, A.~Lopardo, B.~Szymkow, B.~Wagner, E.~Bluemke, J.-M.
  Nounahon, J.~Passerat-Palmbach, K.~Prakash, N.~Rose, et~al.
\newblock Pysyft: A library for easy federated learning.
\newblock In {\em Federated Learning Systems}. 2021.

\end{thebibliography}

\end{document}